\documentclass[twocolumn]{article}
\usepackage{svg}

\usepackage[T1]{fontenc}
\usepackage[utf8]{inputenc}
\usepackage{lmodern}
\usepackage{microtype}

\usepackage[margin=1in]{geometry}
\usepackage{setspace}
\usepackage{parskip}
\usepackage{float}
\usepackage{dblfloatfix}

\usepackage{amsmath, amssymb, amsthm, mathtools}
\usepackage{bm}
\usepackage{bbm}

\usepackage{graphicx}
\usepackage{subcaption}
\usepackage{booktabs}
\usepackage{multirow}
\usepackage{array}
\usepackage{tabularx}
\usepackage{threeparttable}
\usepackage[colorlinks=true, linkcolor=blue, citecolor=blue, urlcolor=blue]{hyperref}
\usepackage{multicol}

\usepackage{booktabs}
\usepackage{tabularx}
\usepackage{threeparttable}
\usepackage{array}
\usepackage[table]{xcolor}
\usepackage{cuted}
\usepackage{caption}
\usepackage[percent]{overpic}
\usepackage{xcolor}

\usepackage{algorithm}
\usepackage{algpseudocode}
\usepackage{listings}
\usepackage{xcolor}

\usepackage{enumitem}
\usepackage{url}
\usepackage{xspace}
\usepackage{comment}

\theoremstyle{definition}

\theoremstyle{remark}

\usepackage{xurl}

\usepackage{caption}
\usepackage{subcaption}
\usepackage{newfloat}
\usepackage[capitalise,noabbrev]{cleveref} 

\Crefname{subfigure}{\textbf{Figure}}{\textbf{Figs.}}

\crefrangelabelformat{subfigure}{%
  #3#1#4--#5\crefstripprefix{#1}{#2}#6%
}

\DeclareCaptionLabelSeparator{bar}{\space|\space}

\DeclareCaptionLabelSeparator{bar}{\space|\space}

\usepackage{multicol}

\DeclareCaptionFormat{twocolcaption}{%
  \begin{minipage}{\linewidth}
  \begin{multicols}{2}
  #1#2#3
  \end{multicols}
  \vspace{0.05em} 
  \end{minipage}
}

\DeclareCaptionLabelSeparator{naturebar}{\space|\space}

\DeclareFloatingEnvironment[
    fileext=loedf,
    name={Extended Data Fig.},
    placement=htbp
]{extendedfigure}

\crefname{extendedfigure}{Extended Data Fig.}{Extended Data Figs.}
\Crefname{extendedfigure}{Extended Data Fig.}{Extended Data Figs.}

\crefname{subextendedfigure}{Extended Data Fig.}{\textbf{Extended Data Figs.}}
\Crefname{subextendedfigure}{Extended Data Fig.}{Extended Data Fig.}

\creflabelformat{subextendedfigure}{#2#1#3}
\crefrangelabelformat{subextendedfigure}{%
  #3#1#4--#5\crefstripprefix{#1}{#2}#6%
}

\usepackage{titlesec}
\titleformat{\subsection}
  {\normalfont\small\bfseries}
  {\thesubsection}{1em}{}

\usepackage[
backend=biber,
style=numeric,
sorting=none
]{biblatex}
\title{\textbf{On the limitations of molecular representations}}
\author{
  Acer Blake \\
  Department of Statistics \\
  University of Oxford \\
  Oxford, UK \\
  \texttt{blake@stats.ox.ac.uk}
    \and
  Charlotte M. Deane \\
  Department of Statistics \\
  University of Oxford \\
  Oxford, UK \\
  \texttt{deane@stats.ox.ac.uk}
  \and
    Fergus Imrie \\
  Department of Statistics \\
  University of Oxford \\
  Oxford, UK \\
  \texttt{imrie@stats.ox.ac.uk}
    \and
}
\date{}

\begin{document}

\twocolumn[
\maketitle

\begin{@twocolumnfalse}
    \begin{abstract}
    \vspace{0.5cm}
     Molecular representations such as fingerprints and learned embeddings play a central role in cheminformatics and small-molecule machine learning. They serve as the primary input for a diverse array of tasks, including clustering, dataset splitting, and virtual screening, yet their evaluation is often limited to downstream model performance on narrow benchmark tasks. We carried out a systematic investigation of the quality of molecular representations, finding that none of the state-of-the-art molecular embedding models capture graph structure, 3D shape, or ADMET endpoint similarity better than fingerprints or physicochemical descriptors.  However, classical fingerprints also revealed their own material limitations. For example, we show that depending on the representation, between 225M and $\sim$4B molecules have non-unique representations in the 10.8B Enamine REAL database, with implications for virtual screening. Our findings provide an insight into the current limitations of molecular representations, outlining future research directions for the development of more robust encodings of small molecules. 
    \end{abstract}

    \vspace{1cm}

\end{@twocolumnfalse}]

Small-molecule representations have been a primary research focus of the cheminformatics community since the inception of the field. Indeed, among the first questions facing a developer of a machine learning model for a task in small-molecule chemistry is what representation should be used as input. While representations abound, three primary input types predominate: string-based notations, molecular graphs, and vector-based encodings \supercite{RepReview1}. 

String-based notations, such as the Simplified Molecular Input Line Entry System (SMILES) \supercite{smiles} format, encode small molecules as one-dimensional sequences of characters wherein the order of atom and bond types represents a particular traversal order of a two-dimensional molecular graph. SMILES is the standard for small-molecule datasets \supercite{RepReview1} and is additionally used as an input representation for many chemical language models (CLMs) \supercite{CLMReview}. Graph-based representations emphasize modelling the connectivity between atoms in two or three-dimensional molecular graphs and include representations such as adjacency matrices. Such representations are commonly used as inputs to graph neural networks (GNNs) \supercite{GNNReview}. Finally, vector-based representations are the primary representation for most small molecule machine-learning applications. Vector-based representations include physicochemical descriptors \supercite{Physchem}, classical molecular fingerprints \supercite{ECFP, TT, AP}, and learned molecular embeddings \supercite{RepReview}. In all three paradigms, the aim is to encode the structural and physicochemical features of molecules as a fixed-length, multi-dimensional vector. 

Fingerprinting algorithms typically encode molecular structures as sparse binary vectors, with common sizes being 1024 or 2048 \supercite{RepReview1}. While fingerprinting algorithms vary with respect to how they encode 2D molecular graphs as a binary array, their high-level operation is the same: encode substructures as integers and hash these integers into an index that is used to mark the presence of the corresponding feature. Implicitly, 0-valued entries of the vector encode the absence of one or many features that map to the associated index. The most commonly used fingerprinting algorithms are extended connectivity fingerprints (ECFPs) \supercite{ECFP}, functional class fingerprints (FCFPs) \supercite{ECFP}, atom pair fingerprints \supercite{AP}, and topological torsion fingerprints \supercite{TT}.  

In contrast to fingerprints, molecular embeddings aim to capture features from molecules that are learned by deep neural networks as part of a supervised \supercite{Minimol} or self-supervised training objective \supercite{MolBERT}. Rather than explicitly crafting algorithms with inbuilt cheminformatic and graph-theoretic heuristics, the objective in molecular representation learning is to select loss functions that lead to an emergent representation of molecular structure and properties. 

Much recent work has focused on benchmarking learned molecular representations against classical descriptors on downstream tasks such as property prediction \supercite{SystemProp, BiogenAndADMETPdv, GNNBench}.  Such studies have revealed that classical representations often match or outperform state-of-the-art molecular embedding models, while maintaining improved interpretability and computational efficiency \supercite{SystemProp, BiogenAndADMETPdv, GNNBench}. Additionally, several studies have investigated the properties of molecular representations more directly, such as whether they capture stereoelectronic properties \supercite{Stereoelec} and activity cliffs \supercite{Actcliffrep}. 

Absent in the literature is a systematic and comprehensive analysis of the quality of learned and classical representations across multiple axes of similarity and application areas.  Consequently, many fundamental questions remain unanswered, such as how often fingerprints fail to uniquely encode structurally distinct molecules in ultra-large libraries, how well molecular representations can distinguish structurally close and distant molecules, to what extent similarity metrics capture three-dimensional shape, how closely representations place molecules with similar ADMET activity, and whether clustering molecules using current representations causes data leakage. We sought to address these questions by assessing a broad selection of representations across three axes of similarity: graph, shape, and chemical. 

Graph similarity pertains to how well a representation captures the two-dimensional topology of a molecule, and whether it can distinguish any two given molecules. Vector-based measures of graph similarity are the foundation underlying ligand-based virtual screening \supercite{RepReview1}, database retrieval \supercite{MolSimRevDB}, clustering \supercite{Butina}, and dataset splitting \supercite{UMAPSplit, LoHi, DataSAIL}. 

Shape similarity relates to the degree of overlap in the three-dimensional conformation of molecules, including the distribution of pharmacophoric properties. Perception of molecular shape has been shown to be advantageous in ligand-based virtual screening \supercite{ShapeScreening}, particularly with respect to scaffold hopping \supercite{ScaffoldHop}. 

Finally, chemical similarity captures the notion that two molecules may have similar activity for a given target or endpoint without necessarily appearing structurally similar by their molecular graphs \supercite{MedchemSim}. In drug discovery, often what is desired is not a measure of shape or graph similarity, but rather a similarity metric that directly facilitates the retrieval of molecules with similar ADMET properties or which bind to the same target. 

Here we evaluate multiple axes of similarity across four fingerprinting algorithms and nine diverse molecular embedding models, spanning CLMs, GNNs, and graph transformers (\textbf{Table} \ref{tab:molecular_representations}). We show that despite being a standard for virtual screening, molecular fingerprints display degeneracy rates over large chemical spaces between 2.2\% and 40\%. Further, we demonstrate that many learned molecular representations fail to capture differences in graph topology, causing them to place highly structurally dissimilar molecules close in representation space, and highly similar molecules far apart. While we found that fingerprints generally better correlate with an NP-hard measure of graph similarity, we show that current ECFP-based clustering practices cause structural leakage, with implications for dataset splitting and the evaluation of machine-learning models trained on molecular data. Indeed, a general observation was that while fingerprints generally outperformed learned representations across each evaluation, neither class of representation can reliably perceive graph, shape or chemical similarity, with downstream consequences for molecular clustering, virtual screening, dataset splitting and property prediction.

\begin{table*}[!p]
\centering
\footnotesize
\setlength{\tabcolsep}{3pt}
\renewcommand{\arraystretch}{1.18}

\definecolor{natureheader}{HTML}{D8D5C2}
\definecolor{naturerow}{HTML}{EFEDE2}
\definecolor{naturerule}{HTML}{6F6F63}
\definecolor{naturetext}{HTML}{111111}

\begin{threeparttable}

\captionsetup{
  labelsep=naturebar,
  labelfont=bf,
  justification=raggedright,
  singlelinecheck=false,
  font=footnotesize
}

\caption{\textbf{Molecular representations used in this study}}
\label{tab:molecular_representations}

\arrayrulecolor{naturerule}
\color{naturetext}
\rowcolors{2}{white}{naturerow}

\begin{tabularx}{\textwidth}{
>{\raggedright\arraybackslash}p{1.9cm}
>{\raggedright\arraybackslash}p{1.8cm}
>{\raggedright\arraybackslash}p{1.4cm}
>{\raggedright\arraybackslash}p{1.95cm}
>{\raggedright\arraybackslash}p{3.75cm}
>{\raggedright\arraybackslash}p{2.25cm}
>{\raggedright\arraybackslash}p{2cm}
}

\toprule


\rowcolor{natureheader}
\multicolumn{7}{l}{
    \textbf{Molecular embedding models}
    \rule[-0.9ex]{0pt}{3.3ex}
} \\

\rowcolor{natureheader}
\textbf{Model} &
\textbf{Model type} &
\textbf{Params} &
\textbf{Train set size} &
\textbf{Pre-training objective} &
\textbf{Loss function} &
\textbf{Stereo-aware} \\
\midrule

MolBERT \supercite{MolBERT} &
CLM (Transformer) &
87M &
1.3M &
Masked language modelling, SMILES equivalence, descriptor prediction &
Multi-task loss &
No \\

\midrule
ChemBERTa \supercite{ChemBERTa} &
CLM (Transformer) &
3.4M &
77M &
Masked language modelling &
Categorical cross-entropy &
No \\

\midrule
MolFormer \supercite{MolFormer} &
CLM (Transformer) &
44.4M &
1.1B &
Masked language modelling &
Multi-class cross-entropy &
No \\

\midrule
MolCLR \supercite{MolCLR} &
GNN &
2.4M &
10M &
Contrastive alignment following atom masking, bond deletion, and subgraph removal &
Scaled cross-entropy contrastive loss &
Yes \\

\midrule
CheMeleon \supercite{CheMeleon} &
GNN &
9.3M &
1M &
Prediction of 1,613 molecular descriptors &
Masked mean squared error &
Yes \\

\midrule
MiniMol \supercite{Minimol} &
GNN &
10.3M &
6M &
Supervised regression and classification &
Weighted MAE, BCE, HCE sum &
No \\

\midrule
UniMol \supercite{UniMol} &
Graph transformer &
47.3M &
209M &
3D position recovery and masked atom prediction &
Smooth L1 and cross-entropy &
Partial \\

\midrule
COATI \supercite{COATI} &
Transformer-GNN hybrid &
12.6M &
140M &
Contrastive alignment of SMILES and 3D molecular point clouds &
Contrastive loss and autoregressive cross-entropy &
No \\

\midrule
MolE \supercite{MolE} &
Graph transformer &
86M &
842M &
Atom environment prediction and supervised fine-tuning on 1,310 ChEMBL readouts &
Multitask binary cross-entropy &
No \\


\midrule

\rowcolor{natureheader}
\multicolumn{7}{l}{
    \textbf{Molecular fingerprints}
    \rule[-0.9ex]{0pt}{3.3ex}
} \\

\rowcolor{natureheader}
\textbf{Fingerprint} &
\textbf{Type} &
\textbf{Size} &
\textbf{Environment} &
\textbf{Information encoded} &
\textbf{Implementation} &
\textbf{Stereo-aware} \\
\midrule

Atom Pair \supercite{AP}&
Topological &
1024 bits &
Linear paths up to 30 bonds &
Pairs of atom environments and their topological separation &
RDKit atom-pair fingerprint &
Optional \\

Topological Torsion \supercite{TT}&
Topological &
1024 bits &
Linear 4-atom sequences &
Quadruplets of four linearly bonded atoms and their atom environments &
RDKit topological torsion fingerprint &
Optional \\

\midrule

\midrule
ECFP4 \supercite{ECFP}&
Circular &
1024 bits &
Radius 2 substructures&
Local atom-centred environments based on atomic connectivity &
RDKit Morgan fingerprint &
Optional \\

\midrule
FCFP4 \supercite{ECFP}&
Circular &
1024 bits &
Radius 2 substructures&
Local atom-centred environments based on pharmacophoric features &
RDKit Morgan fingerprint with feature invariants &
Optional \\


\midrule

\rowcolor{natureheader}
\multicolumn{7}{l}{
    \textbf{Physicochemical descriptors}
    \rule[-0.9ex]{0pt}{3.3ex}
} \\

\rowcolor{natureheader}
\textbf{Descriptor} &
\textbf{Type} &
\textbf{Features} &
\textbf{Input} &
\textbf{Information encoded} &
\textbf{Implementation} &
\textbf{Stereo-aware} \\
\midrule

RDKit \supercite{rdkit_calcmoldescriptors}&
Physico-chemical descriptor vector &
217 & 
2D molecular graph &
Topology, charge-related properties, surface-area-related properties, and structural features &
RDKit {\scriptsize\texttt{CalcMolDescriptors}} &
No \\

\bottomrule

\end{tabularx}

\arrayrulecolor{black}

\caption*{\footnotesize
“Stereo-aware” indicates whether stereochemical information is explicitly encoded in the molecular representation or made available during pre-training. For fingerprint representations, "Environment" refers to the structural environment encoded by each fingerprint feature. The number following ECFP and FCFP fingerprints is equal to twice the radius set as a parameter.}

\end{threeparttable}
\end{table*}

\section*{Results}

\subsection*{Fingerprints are degenerate in ultra-large libraries}

Due to their fixed-length and discrete nature, molecular fingerprints display degeneracy, a property wherein structurally distinct molecules map to identical representations. Degeneracy has been reported in previous literature on fingerprints \supercite{EarlyDegen, CountYourBits}. However, little follow-up has been conducted on the scale of this phenomenon. To date, the largest analysis of fingerprint degeneracy examined only one fingerprint type for a sample of 1.4 million molecules \supercite{NeuralDecipher}. However, virtual screening efforts are now routinely conducted over ultra-large libraries \supercite{ULConfDocking, ULDeepDocking}, with molecules numbering in the tens of billions. Such `active learning' or `deep docking' studies compute billions of fingerprints to accelerate structure-based approaches \supercite{ULDeepDocking, ULFCFPDocking}. Absent in the literature is a systematic examination of the scale of fingerprint degeneracy in real-world virtual-screening libraries across fingerprint types, as well as a taxonomy of real molecular substitutions made by medicinal chemists that may lead to degenerate representations. 

To address these questions, we generated fingerprints for the entirety of the Q4 2025 Enamine REAL database of 10.8 billion molecules.  After pre-processing (\textbf{Methods}), 10.3 billion molecules remained. Each fingerprint was generated for every molecule twice: once with a chirality flag that enforces the inclusion of stereochemical information and once without. 

\begin{figure*}[p]
    \rule{\linewidth}{1.2pt}
    \centering

    \includegraphics[scale=0.4]{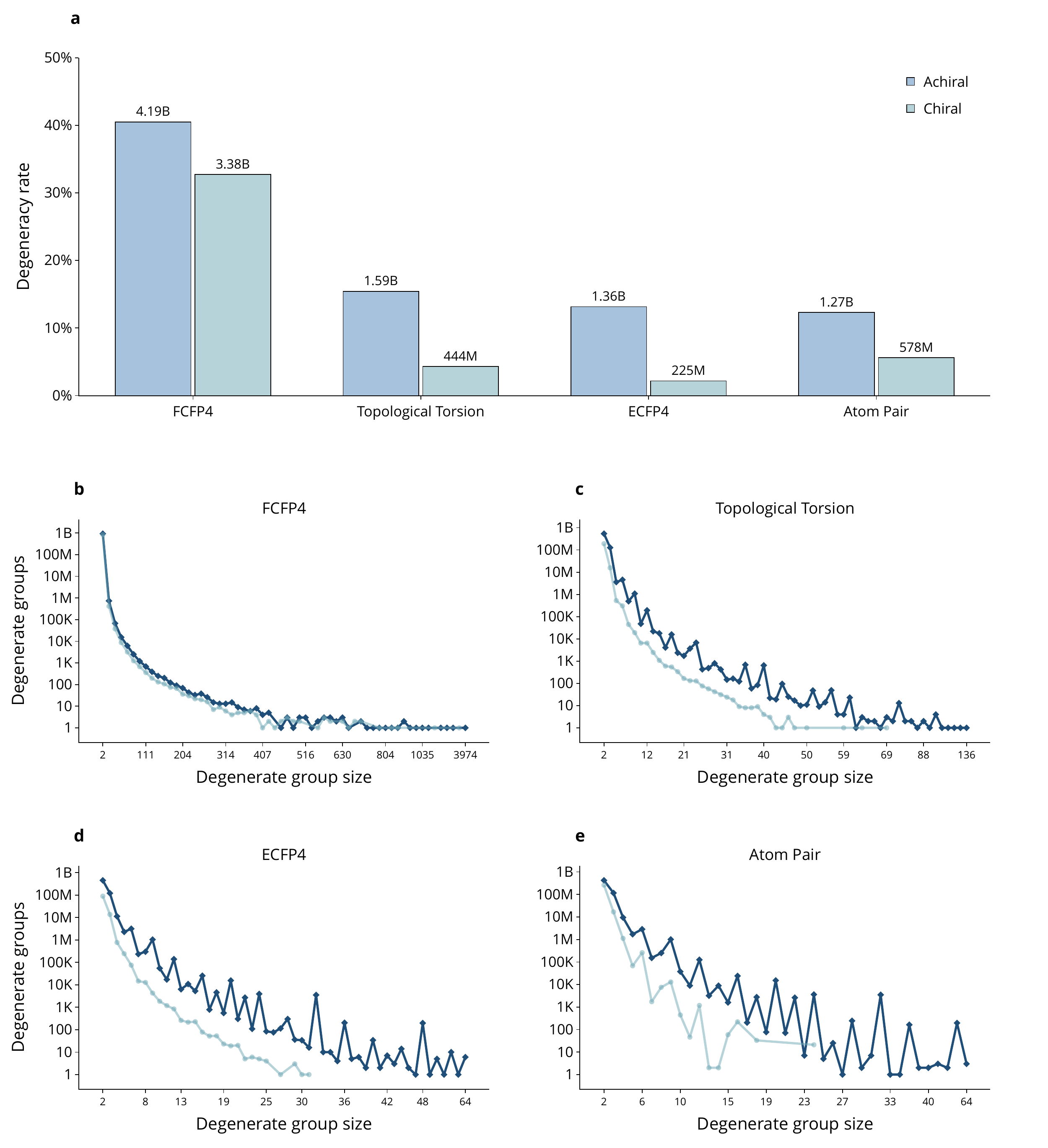}

    \caption{\textbf{Degeneracy rates for molecular fingerprints across the filtered 10.3 billion molecule Enamine REAL space}. 
    \textbf{(a)} Bar chart showing the degeneracy rates for each of the four tested fingerprint types. A molecule is considered to be degenerate if it participates in a group of two or more non-identical molecules that map to the same fingerprint. \textbf{(b-e)} Group sizes for sets of degenerate molecules across FCFP4, topological torsion, ECFP4, and atom pair fingerprints, respectively. Parameter settings for the fingerprints are given in Table \ref{tab:molecular_representations}.}

    \rule{\linewidth}{1.2pt}
    \label{fig:degeneracy}
    \phantomsubcaption\label{fig:degeneracy-a}
    \phantomsubcaption\label{fig:degeneracy-b}
    \phantomsubcaption\label{fig:degeneracy-c}
    \phantomsubcaption\label{fig:degeneracy-d}
    \phantomsubcaption\label{fig:degeneracy-e}
\end{figure*}

Our analysis revealed that FCFP4 fingerprints were the most susceptible to degeneracy, with 40.5\% and 32.7\% of molecules being degenerate under this fingerprint for the achiral and chiral versions, respectively (\textbf{Fig.} \ref{fig:degeneracy-a}). This was followed by topological torsion (achiral: 15.4\%, chiral: 4.3\%), atom pair (achiral: 12.3\%, chiral: 5.6\%), and finally ECFP4 (achiral: 13.2\%, chiral: 2.2\%). Degenerate molecules include stereoisomers that neither parameterization of the fingerprints can distinguish \supercite{Stereo}, as well as structurally distinct molecules. We find the rates of degeneracy to be surprisingly high, with even the lowest rate of 2.2\% for chiral ECFP4 to be equivalent to 225 million molecules with at least one structurally distinct partner yielding an identical fingerprint. 

An inverse relationship was observed between the size of degenerate groups of molecules and their frequency (\crefrange{fig:degeneracy-b}{fig:degeneracy-e}), with maximum degenerate group sizes of 3,974 (chiral: 2,196), 64 (chiral: 31), 64 (chiral: 24) and 136 (chiral: 69) for FCFP4, ECFP4, atom pair and topological torsion fingerprints, respectively. Group sizes are greater for achiral fingerprints due to the high number of stereoisomers that differ only by toggling stereobonds, which the chirality-aware fingerprints are able to distinguish more frequently.  

Both ECFPs and FCFPs have been used in ultra-large-scale virtual screening of the Enamine REAL space, often with chirality disabled \supercite{ULConfDocking, ULFCFPDocking, EnamineREALDock}. For example, one recent study screened 37 billion molecules from the Enamine REAL space using achiral FCFPs\supercite{ULFCFPDocking}. Despite the successes of these studies in reporting hits, many hundreds of millions to billions of molecules necessarily received identical predictions despite being structurally and physicochemically distinct, conflating not just differences in affinity, but also binders and non-binders.

\begin{figure*}[p]
    \centering
    \rule{\linewidth}{1.2pt}

    \includegraphics[width=1.0\linewidth]{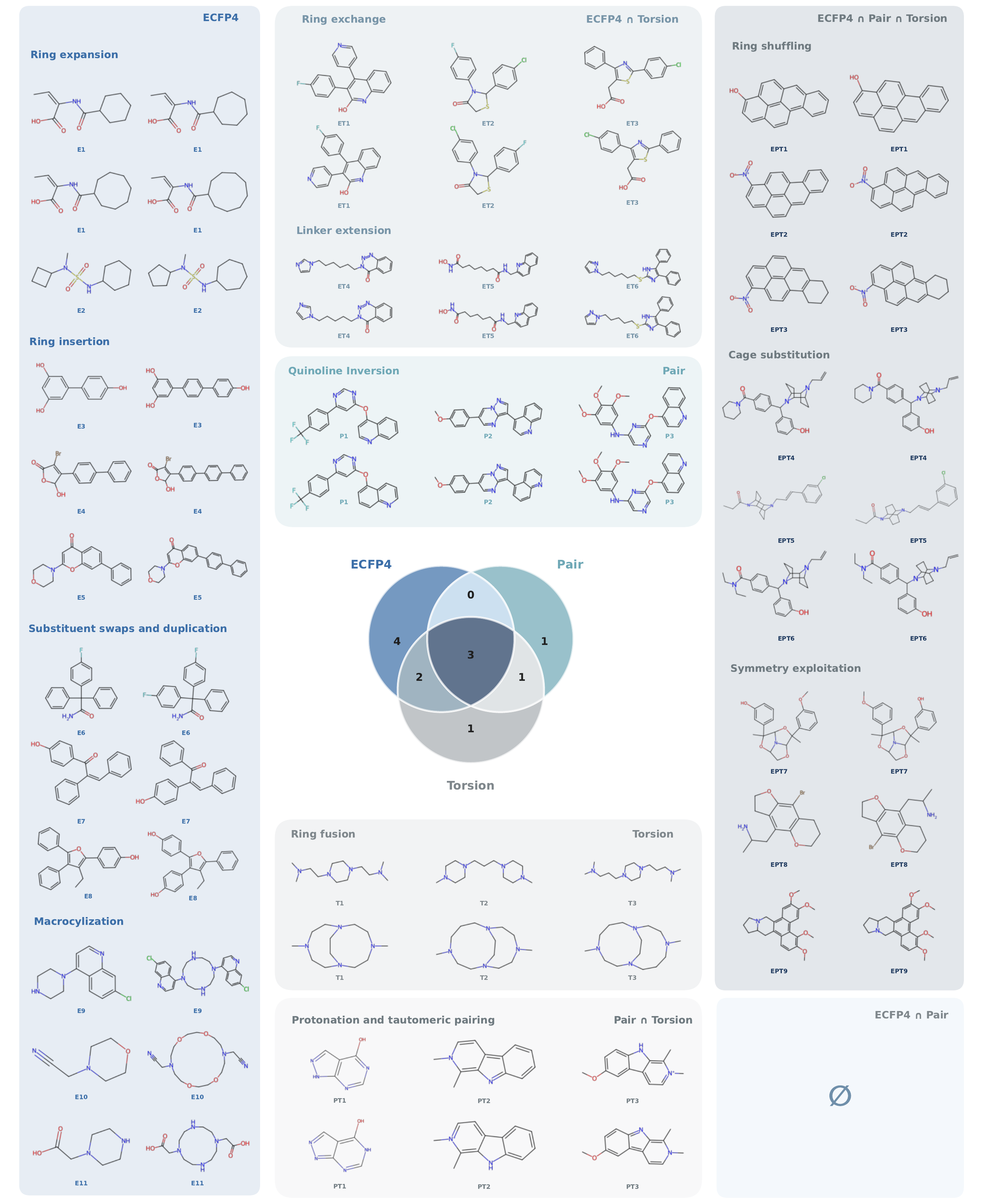}

    \caption{\textbf{Degeneracy failure modes across three fingerprint types}. The figure depicts a non-exhaustive set of transformations that can be applied to preserve fingerprint identity across ECFP4, atom pair, topological torsion fingerprints, and every intersection between them. Each letter label denotes the fingerprint type(s) for which degeneracy holds, while molecules with the same number have identical fingerprints. All examples are sourced from Enamine and ChEMBL. The Venn diagram at center indicates how many consistent and repeatable failure modes were found for each fingerprint type and combination.}
    \label{fig:degeneracy-failure-modes}

    \rule{\linewidth}{1.2pt}
\end{figure*}

We provide a hand-crafted taxonomy of the core failure modes we observed in the sets of degenerate molecules that exist in ChEMBL \supercite{ChEMBL} and Enamine. \textbf{Fig.} \ref{fig:degeneracy-failure-modes} highlights these for ECFP4, atom pair, and topological torsion fingerprints. Among the most prominent failure modes were carbon chain elongation, carbon chain shifting, ring expansion, substructure duplication, and substituent relocation. While less frequent, \textbf{Fig.} \ref{fig:degeneracy-failure-modes} shows that it is possible for all three fingerprints to be degenerate for the same molecular pairs. Thus, simply ensembling fingerprints in the hopes that this will generate enough features to distinguish all molecular pairs is not a panacea for degeneracy.

\begin{figure*}
    \centering
    \hspace{-0.3cm}
    \includegraphics[width=1.0\textwidth]{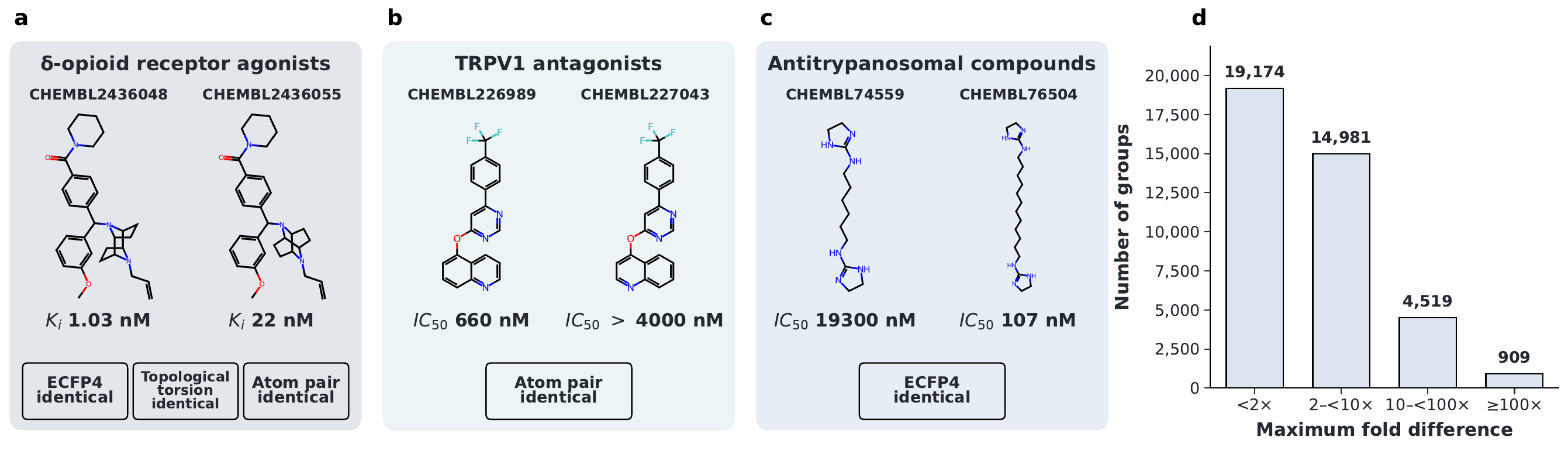}
    \captionof{figure}{\textbf{Activity differences for degenerate molecules in ChEMBL}. \textbf{(a)} $\delta$-opioid receptor agonists indistinguishable by all fingerprints \supercite{diaza}, \textbf{(b)} Amgen TRPV1 antagonists indistinguishable by atom pair fingerprints \supercite{amgen}. \textbf{(c)} Antitrypanosomals indistinguishable under ECFP4 \supercite{antitrypanosomal}. Each example pair is published within one paper and assayed under the same experimental conditions. (d) Number of chiral, 1024-bit ECFP4 degenerate groups in the same ChEMBL assay by maximum within-group fold change.}
    \label{fig:matched-pairs}
    \phantomsubcaption\label{fig:matched-pairs-a}
    \phantomsubcaption\label{fig:matched-pairs-b}
    \phantomsubcaption\label{fig:matched-pairs-c}
    \phantomsubcaption\label{fig:matched-pairs-d}
      
\end{figure*}

We provide constructions demonstrating how several of these modifications can be made to arbitrary molecules satisfying certain criteria to create degenerate analogues in \textbf{Supplementary Figs.} \textbf{1-8}. As these examples are sourced from ChEMBL and Enamine, these constructions show that fingerprints are insensitive to molecular modifications that medicinal chemists regularly make. Further, these modifications can lead to material differences in activity. We show three examples from ChEMBL in \textbf{Fig.} \ref{fig:matched-pairs}, where even simple changes in linker length can result in 180-fold changes in potency. Across approximately 40,000 groups of ECFP4-degenerate molecules with measurements in the same assay, approximately half showed a maximum activity difference of at least two-fold, and more than 5,000 displayed greater than a ten-fold change (\textbf{Fig. \ref{fig:matched-pairs-d}}). 

\subsection*{No representation captures topology robustly}

A primary objective of molecular representations is to capture the similarity between molecular graphs. It has previously been shown that Tanimoto similarity may provide radically different estimations of structural similarity between molecular pairs when compared with human intuition \supercite{BitStringSim}. Moreover, a recent study by Kretschmer et al \supercite{MCES} provided examples of the difference between Tanimoto similarity and the weighted maximum common edge substructure (MCES) distance, an NP-hard graph similarity metric. We modified this metric for use as a measure of ground-truth graph similarity (\textbf{Methods}).

Denoted $\mathrm{d_{MCES^w_{norm}}}\in [0, 1]$, our metric indicates the weighted cost of deleting bonds until two molecules are reduced to a shared bond structure, normalized by the worst case of deleting all bonds. \textbf{Extended Data Fig.} \ref{fig:mces-hexbin-a} provides an example calculation of the metric for two molecules. After calculating pairwise $\mathrm{d_{MCES^w_{norm}}}$ values using the MCES dataset (\textbf{Methods}) computed by Kretschmer et al \supercite{MCES}, we quantified the Spearman correlation with Tanimoto and cosine similarities (\crefrange{fig:mces-hexbin-b}{fig:mces-hexbin-m}) over the representations given in \textbf{Table} \ref{tab:molecular_representations}.

\begin{figure*}[p]
    \rule{\linewidth}{1.2pt}
    \centering

    \includegraphics[scale=0.475]{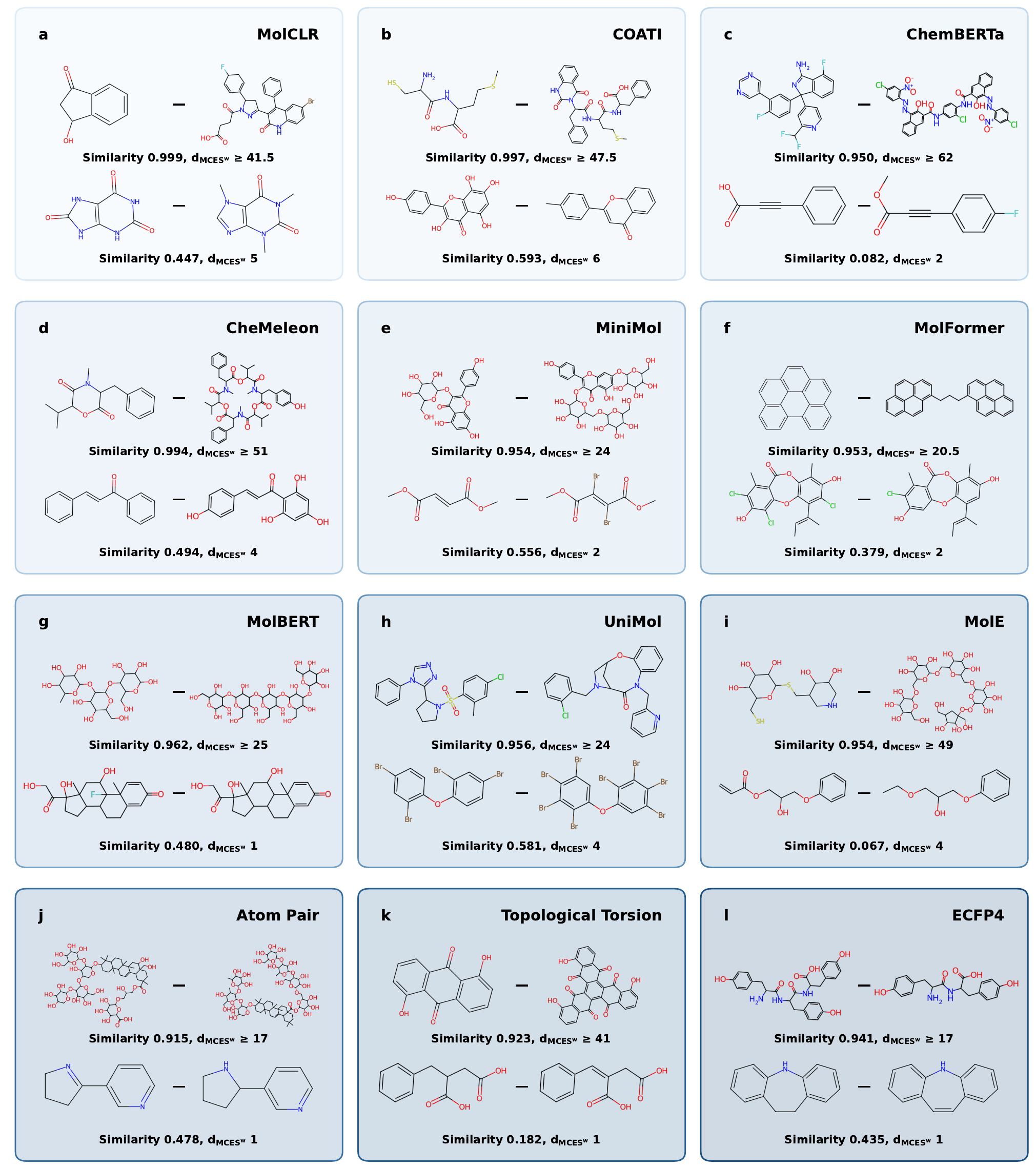}


    \vspace{0.4em}

    \caption{\textbf{Disagreement between MCES and similarities defined over molecular representations}.
    \textbf{(a--l)} Poor calibration of cosine similarities
    \textbf{(a--i)} and Tanimoto similarities \textbf{(j--l)}
    with respect to global molecular graph topology.
    Examples for each model are divided into two: the top molecular
    pair shows high representation similarity and high MCES distance,
    while the bottom pair shows low relative representation similarity and low
    MCES distance. Due to their extreme cosine similarity distributions, examples for the MolCLR \textbf{(a)} and COATI
    \textbf{(b)} models were rescaled using min-max percentile scaling.}

    \label{fig:mces-examples}
    \phantomsubcaption\label{fig:mces-examples-a}
    \phantomsubcaption\label{fig:mces-examples-b}
    \phantomsubcaption\label{fig:mces-examples-c}
    \phantomsubcaption\label{fig:mces-examples-d}
    \phantomsubcaption\label{fig:mces-examples-e}
    \phantomsubcaption\label{fig:mces-examples-f}
    \phantomsubcaption\label{fig:mces-examples-g}
    \phantomsubcaption\label{fig:mces-examples-h}
    \phantomsubcaption\label{fig:mces-examples-i}
    \phantomsubcaption\label{fig:mces-examples-j}
    \phantomsubcaption\label{fig:mces-examples-k}
    \phantomsubcaption\label{fig:mces-examples-l}

    \rule{\linewidth}{1.2pt}
\end{figure*}

Atom pair fingerprints were the representation most strongly correlated with graph topology, with a Spearman correlation of -0.82, the next closest being topological torsion fingerprints with a Spearman correlation of -0.73. The ECFP4 representation was significantly less correlated with global graph shape, with a Spearman correlation of -0.63. MolFormer was the learned representation most correlated with global graph topology, with a correlation of -0.72. However, 5 of 9 learned molecular representations displayed correlations greater than -0.5, indicating weak alignment with global graph topology. Indeed, with a Spearman correlation of 0.1, MolE cosine similarity proved to be effectively uncorrelated with $\mathrm{d_{MCES^w_{norm}}}$. Marginal plots for the representations also reveal a difference in distribution shape between molecular fingerprints and learned embeddings. Tanimoto similarities for the MCES dataset are right-skewed, while all of the learned representations are left-skewed, bimodal, collapsed, or normal, indicating a greater tendency toward ranking molecules as similar.

\textbf{Fig.} \ref{fig:mces-examples} provides examples of poor calibration of cosine and Tanimoto similarities with respect to the unnormalized weighted MCES distance, $d_{MCES}^w$, for each of the 12 molecular representations. For the classical representations (\crefrange{fig:mces-examples-j}{fig:mces-examples-l}), these examples highlight the key flaw of binary fingerprints: repeated occurrence of the same substructures in one molecule with respect to a reference allows two molecules of very different sizes to appear highly similar. We find that across learned molecular representations (\crefrange{fig:mces-examples-a}{fig:mces-examples-i}), molecules with highly dissimilar scaffolds are ranked as considerably more similar than molecules which differ by a handful of bonds. For example, \textbf{Fig.} \ref{fig:mces-examples-d}  shows that CheMeleon may assign a cosine similarity of 0.994 to two molecules that differ in graph structure by a $d_{MCES}^w$ of more than 50, while two near-identical molecules with a distance of 4 are assigned a cosine similarity of 0.494. This behaviour suggests that current learned representations do not capture graph shape robustly, and for applications in which perception of graph shape is important, such as ligand-based virtual screening, these models are unlikely to be as suitable as classical fingerprints.

\begin{figure*}[p]
        \rule{\linewidth}{1.2pt}
        \centering
        \hspace{-0.4cm}
        \includegraphics[scale=0.2]{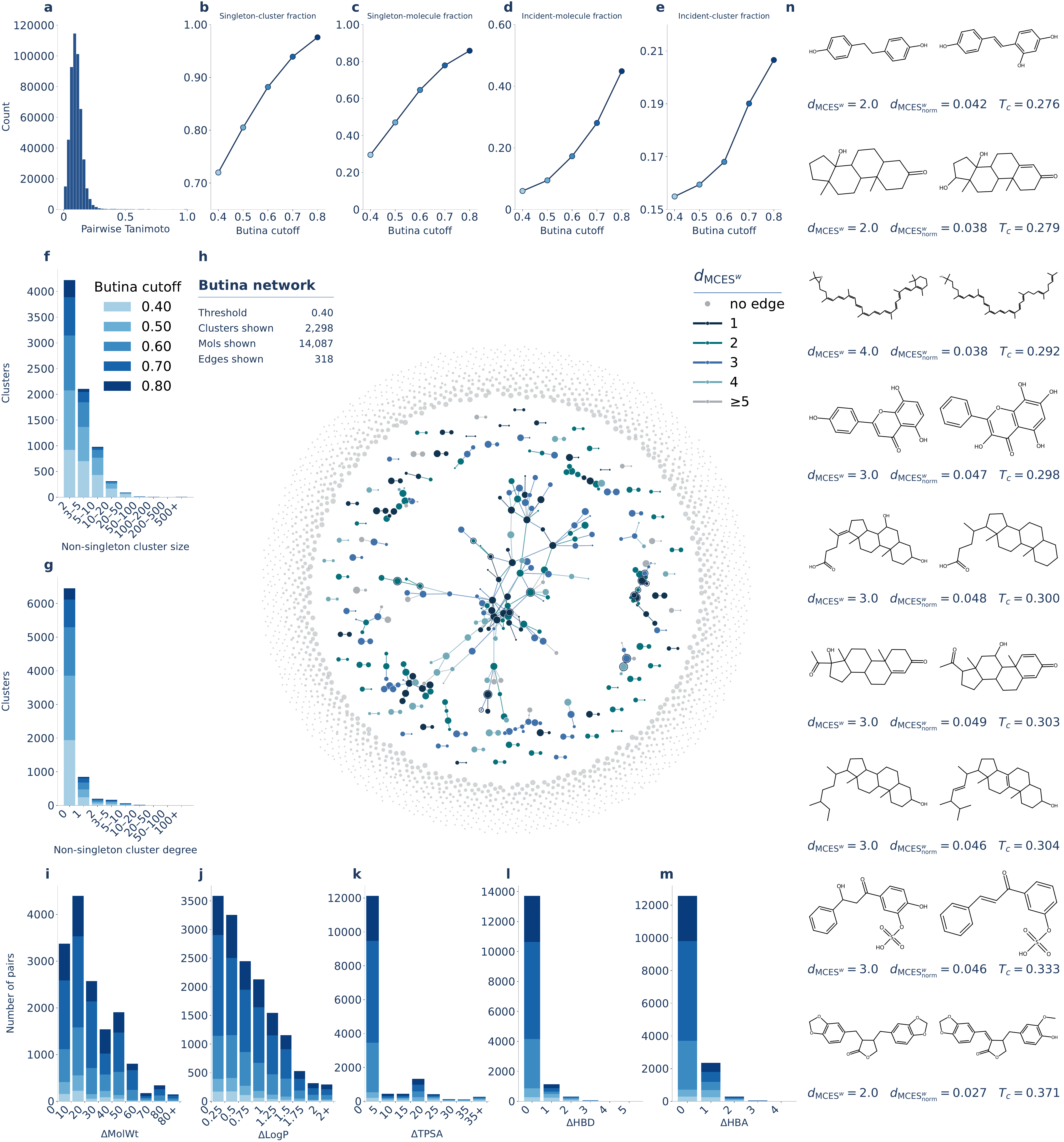}

    \caption{\textbf{Butina cluster leakage}. \textbf{(a)} Pairwise Tanimoto similarity distributions for the MCES dataset. \textbf{(b)} Fraction of singleton clusters across Butina thresholds. \textbf{(c)} Fraction of the molecules contained in singleton clusters. \textbf{(d)} Fraction of molecules with $d_{MCES^w_{norm}} < 0.05$ and $d_{MCES^w}$ $< 10$ to other molecules in different clusters (edges). \textbf{(e)} Fraction of clusters with at least one edge incident to another cluster using the same threshold as in \textbf{(d)}. \textbf{(f)} Stacked cluster size distributions across a selected Butina Tanimoto threshold range. \textbf{(g)} Stacked cluster degrees across different thresholds. \textbf{(h)} Network showing 2,298 clusters using a Butina threshold of 0.4. Each edge represents a pair of molecules assigned to different clusters with $d_{MCES^w_{norm}} < 0.05$ and $d_{MCES^w}$ $< 10$. Nodes are scaled by molecule count and coloured according to minimum incident $d_{MCES^w}$. \textbf{(i-m)} Stacked differences for each MCES pair across physicochemical properties. \textbf{(n)} Example molecular pairs from \textbf{(h)}. Each pair contains two molecules from different clusters with low MCES distance. $T_c$ denotes the Tanimoto similarity coefficient. }
    \rule{\linewidth}{1.2pt}
    \label{fig:butina}
    \phantomsubcaption\label{fig:butina-a}
    \phantomsubcaption\label{fig:butina-b}
    \phantomsubcaption\label{fig:butina-c}
    \phantomsubcaption\label{fig:butina-d}
    \phantomsubcaption\label{fig:butina-e}
    \phantomsubcaption\label{fig:butina-f}
    \phantomsubcaption\label{fig:butina-g}
    \phantomsubcaption\label{fig:butina-h}
    \phantomsubcaption\label{fig:butina-i}
    \phantomsubcaption\label{fig:butina-j}
    \phantomsubcaption\label{fig:butina-k}
    \phantomsubcaption\label{fig:butina-l}
    \phantomsubcaption\label{fig:butina-m}
    \phantomsubcaption\label{fig:butina-n}
\end{figure*}

\subsection*{Clustering ECFP4 fingerprints leads to leakage}

A common task in  small-molecule machine learning is clustering, a process which entails assigning structurally related molecules to groups, typically on the basis of their molecular fingerprints \supercite{Butina, LoHi, DataSAIL}. The most common application of this is in dataset splitting, wherein we desire to assign distinct clusters of molecules entirely to train or test, so as to minimise structural data leakage. 

The most popular clustering method in small-molecule machine learning is the Butina algorithm \supercite{Butina}, a greedy optimization algorithm that iteratively creates clusters from the molecule with the greatest number of unassigned neighbours with distances under a given threshold. While the Butina algorithm can operate over any distance matrix, a widespread practice is to use Tanimoto similarities over ECFP4 fingerprints \supercite{cutoff-example1, cutoff-example2}. We therefore focused our analysis on this representation.

To investigate whether Butina clustering using ECFP4 Tanimoto similarity leads to structural data leakage, we clustered $\sim 20$ thousand molecules from the MCES dataset using a range of Tanimoto thresholds from 0.4 to 0.8. Sampled from a range of molecular databases and datasets, the MCES dataset is ECFP4 chemically diverse, as shown in \textbf{Fig.} \ref{fig:butina-a}. 

Our analysis revealed that at both high and low similarity thresholds, distinct clusters often contain highly structurally similar molecules. \textbf{Fig.} \ref{fig:butina-h} shows an example network of close MCES neighbours split across different clusters. The example uses a conservative Tanimoto similarity threshold of 0.4, which is widely used as a cutoff for molecular novelty \supercite{LoHi, sim-example} and Butina clustering \supercite{cutoff-example1, cutoff-example2}. Each node represents a non-singleton cluster, while edges indicate that at least one pair of molecules split between the two incident clusters are highly similar. A perfect clustering would show a completely disconnected network, indicating no examples of highly structurally similar molecules split between different clusters, whereas the example shows over 300 edges. \textbf{Fig.} \ref{fig:butina-n} highlights several examples of highly-similar molecules that are split between distinct clusters by the algorithm.   Due to the moderate correlation of ECFP4 Tanimoto similarity with global graph structure, it is not possible to use Tanimoto similarity to diagnose the issue. For example, each pair in \textbf{Fig.} \ref{fig:butina-n} has a similarity below the threshold of 0.4, with one clearly highly structurally similar pair having a similarity of 0.276. 

As expected, we found that these structure-similar analogues also have similar physicochemical properties. \crefrange{fig:butina-i}{fig:butina-m} show the distribution of differences in molecular weight, logP, TPSA, and hydrogen bond donors and acceptors for the pairs. For each distance threshold, the distribution of differences in the properties of close analogues is modest.

While the number of these pairs decreases with a reduction in the similarity threshold used, this comes at the expense of greatly increasing the cluster sizes and heterogeneity. \crefrange{fig:butina-b}{fig:butina-e} show the effects on the fraction of singleton clusters, molecule membership to singleton clusters, molecule edge incidence, and cluster edge incidence as the cluster threshold varies.  

Distinct from known cluster quality limitations of the Butina algorithm \supercite{NCICluster, UMAPSplit}, these findings partially explain the observation that models trained on datasets split using the Butina algorithm tend to overestimate virtual screening performance \supercite{UMAPSplit}. If Butina splitting is used for train test splits, many structurally similar molecules with near-identical physicochemical properties will be split between train and test sets. We note that many other splitting methods are reliant on clustering fingerprints, such as the UMAP \supercite{UMAPSplit} split, DataSAIL \supercite{DataSAIL} , BitBIRCH \supercite{BitBIRCH} and Lo-Hi \supercite{LoHi}. It is likely that regardless of optimization objective, all of these methods will suffer the same failure mode as Butina splitting when used on current representations.

\subsection*{Fingerprints track 3D shape more consistently}

To assess the ability of molecular representations to capture three-dimensional shape, we investigated the extent to which 2D similarity scores capture shape and pharmacophoric (color) overlap (\textbf{Methods}).

The results broadly recapitulate the findings of the graph similarity experiments: with the exception of COATI in the case of shape score, fingerprints correlate more strongly with shape and color scores across their similarity range than every embedding (\textbf{Extended Data Fig.} \ref{fig:shape-evaluations-a}). Despite correlating poorly with 2D graph topology, the performance of COATI is explained by its contrastive learning objective focused on aligning SMILES strings and 3D point clouds. However, COATI does not show material improvement in correlation strength for shape score (Spearman = 0.80) over atom pair fingerprints (Spearman = 0.79), and displays weaker correlation with color score than all fingerprints. 

The extent to which representations correlate with shape and color is highly dependent upon the sampling strategy. Using the rolling Spearman correlation within bins of molecular pairs, \crefrange{fig:shape-evaluations-b}{fig:shape-evaluations-m} show that within narrow bands of similarity, none of the representations correlate strongly with shape and color. Rather, it is the general increase in shape and color scores across broad ranges of similarity that contributes to the apparent associations. Thus, use of current representations to rank close analogues for 3D similarity is unlikely to perform well, as suggested by recent work \supercite{SAND}. 

\subsection*{Descriptors better capture ADMET similarity}

To evaluate the degree to which molecular representations place molecules with similar activity close together in representation space, we used the Biogen ADMET dataset, a high-quality industrial campaign where all assays were conducted within the same lab\supercite{Biogen}. We excluded two representations from our analyses, namely MiniMol and MolE, as they have been trained on molecules with similar or identical structures to those in the Biogen dataset for related or even identical endpoints. Additionally, we added the RDKit physicochemical descriptor vector (PDV) containing 217 features as a further benchmark. For these 11 representations, we considered the median endpoint difference between molecules as the top closest molecular pairs to consider decreased.  

All representations show a decrease in average endpoint difference as the number of top-most similar pairs to consider decreases, suggesting that all representations are at least somewhat useful in discriminating properties relevant to ADMET endpoint similarity (\crefrange{fig:chem-evals-a}{fig:chem-evals-d}). While there is variability in the trajectories across the similarity thresholds, we observed that for all four of the endpoints, a simple RDKit physicochemical descriptor vector baseline produced the lowest difference in endpoint value for the top 1,000 pairs. This suggests that learned molecular representations are no better at placing molecules with similar endpoint values closer in representation space than standard cheminformatic descriptors. Indeed, Jaccard similarity matrices show that there is generally low agreement between molecular representations with respect to the most similar molecules (\crefrange{fig:chem-evals-e}{fig:chem-evals-h}).

We conducted a follow-up analysis assessing the association between property landscape roughness and cross-validated model performance on the Biogen dataset. Expanding on previous findings \supercite{ROGIXD, BiogenAndADMETPdv}, we found that the RDKit PDV induces the smoothest landscapes for 3 of 4 tested endpoints, and that this tracks with lower median RMSE across models for those same endpoints (\crefrange{fig:rogi-evals-a}{fig:rogi-evals-c}).

\section*{Discussion}

Used at almost every stage in the computational drug discovery pipeline, vector-based molecular representations play a central role in small-molecule machine learning. In this work, we demonstrate that both classical molecular fingerprints and learned molecular embeddings suffer from consistent failure modes with respect to their suitability for assessing similarity between pairs of molecules. Further, we show that failure to capture molecular similarity via these representations necessarily has a downstream impact on a broad range of applications in small-molecule machine learning, including virtual screening, clustering, and property prediction. 

Our degeneracy analyses over the 10.8B Enamine REAL database revealed that hundreds of millions to billions of structurally distinct molecules are indistinguishable under common fingerprinting algorithms. This has implications for virtual screening of ultra-large libraries; if our models cannot distinguish molecules by their input features, it is not possible for them to predict distinct labels, such as binding affinity or docking scores. 

Examination of the failure modes that lead to degeneracy revealed that modifications  medicinal chemists regularly make are often imperceptible to all of the most commonly used fingerprinting algorithms. While degeneracy rates would be lower with count-based fingerprints \supercite{CountYourBits}, it has been shown that degenerate representations can also be constructed for these fingerprints using combinatorial enumeration \supercite{DeterministicEnum}.

Beyond degeneracy, we have shown that both learned and classical molecular representations struggle to capture global graph topology. Cosine similarity between embeddings generated by many state-of-the-art learned molecular representation models correlated poorly with the normalized maximum common edge subgraph measure of global graph similarity, revealing that such models are not calibrated to capture the bond structure of molecules. Similarly, while fingerprint Tanimoto similarities generally correlated more strongly with global graph topology, examples abound of unintuitive similarity values generated by these algorithms. 

Our findings reveal the uneven nature of the capabilities of molecular representations, highlighting the necessity of a new wave of small molecule encodings, as well as evaluation protocols not solely predicated on downstream predictive performance.

\section*{Methods}

\subsection*{Enamine degeneracy analysis}

The 10.8B Enamine REAL database was used to conduct degeneracy analyses for ECFP4, FCFP4, atom pair and topological torsion fingerprints. Files were downloaded from the Enamine website\supercite{enamine}. All molecules in the REAL database are compliant with Lipinski's rule of five and Veber criteria: molecular weight $\leq 500$, SlogP $\leq 5$, hydrogen bond acceptors $\leq 10$, hydrogen bond donors $\leq 5$, rotatable bonds $\leq 10$, and topological polar surface area $\leq 140$. As provided, the REAL database was divided into 10 files, with molecules separated into heavy atom count intervals. Each file contains approximately 1B SMILES strings. SMILES strings are given in a CXSMILES format, which allows for the specification of additional molecular properties through the inclusion of metadata at the end of each string. One such feature is enhanced stereochemistry. Due to the inability of RDKit fingerprinting methods to utilize enhanced stereochemistry, we observed that these additional molecular annotations may lead to the inclusion of the same SMILES string multiple times in each file, trivially increasing the degeneracy rate. We therefore stripped all CXSMILES annotations and deduplicated the resulting SMILES strings within each file. This led to approximately 532 million molecules being removed from the dataset.

Naively calculating a pairwise Tanimoto similarity matrix over the entire pre-processed 10.3B REAL database would be computationally intractable, as the resulting matrix would contain $\sim10^{20}$ entries. Thus, to calculate degeneracy rates for each fingerprint type, we employed a four-stage process. Given a fingerprint type, we first generated fingerprints for the entire REAL database. We then hashed all fingerprints into one of 1,024 possible files. By selecting a hash function that approximately uniformly distributes the fingerprints, we ensured that each file contained roughly 10M fingerprints. Critically, given a deterministic hash function, all identical fingerprints are guaranteed to be in the same file. The final stages were to sort each file, count adjacent runs of identical fingerprints, and sum these across files to calculate the total degeneracy rate.

Fingerprints were generated using RDKit via the rdFingerprintGenerator utility within the Chem module. All fingerprints were generated with a hashed vector length of 1024. ECFP and FCFP fingerprints were set with radius 2, while atom pair and topological torsion fingerprints were generated with default parameters. 

\subsection*{ChEMBL degeneracy analysis}

We conducted our degeneracy analysis of ChEMBL using the ChEMBL37 SQLite download \supercite{chembl37}. We extracted activity data only from assays with a non-empty journal field and a listed doi, and only compared the activity data of molecules listed within the same assay. Where a molecule had multiple readouts for the same assay, the median value was taken as representative. We excluded measurements with non-equality quantifiers, such as IC$_{50} < 20$nM, and only retained activites reported with an equality relation. We did not de-duplicate degenerate groups. Thus, the counts given in \textbf{Fig.} \ref{fig:matched-pairs-d} include cases in which a deposition associated with a paper performed multiple distinct assays for the same degenerate molecules. Similarly, distinct depositions may have assayed the same degenerate molecules under a different set of experimental conditions or for a different assay. These cases too are retained in the counts. 1024 bit ECFP4 fingerprints with chirality enabled were used.

\subsection*{Graph similarity}

We evaluated graph similarity using the weighted maximum common edge subgraph (MCES) distance as defined by Kretschmer et al \supercite{MCES}. Given two graphs \(G_1 = (V_1, E_1)\) and \(G_2 = (V_2, E_2)\), their \emph{Maximum Common Edge Subgraph} (MCES) is a graph
\(G_c = (V_c, E_c)\), where
\[
V_c \subseteq V_i, \qquad E_c \subseteq E_i, \qquad i \in \{1,2\},
\]
such that \(|E_c|\) is maximal. We define the MCES distance between the two
graphs as
\[
d_{\mathrm{MCES}}(G_1, G_2)
=
|E_1| + |E_2| - 2|E_c|.
\]

Qualitatively, this can be thought of as the number of bonds that must be removed from both molecules to make them isomorphic. In particular, Kretschmer et al use a weighted version of the MCES, wherein weights are assigned to bonds to represent the magnitude of their contribution to the overall weight of the molecular graph. Single bonds are assigned a weight of 1, aromatic bonds a weight of 1.5, double bonds a weight of 2, and triple bonds a weight of 3. Mathematically, let $G_1$ and $G_2$ be molecular graphs such that
\[
M_i = (G_i, a_i, b_i), \qquad i \in \{1,2\},
\]
where \(a_i : V_i \to \Sigma\) denotes the atom type and
\(b_i : E_i \to \{1, 1.5, 2, 3\}\) denotes the bond type. The weighted MCES distance is

\[
\resizebox{\columnwidth}{!}{$\displaystyle
d_{\mathrm{MCES}^w}(M_1,M_2)
=
\sum_{e\in E_1} b_1(e)
+
\sum_{e\in E_2} b_2(e)
-
2\sum_{e\in E_c}\min\{b_1(e),b_2(e)\}
$}
\]

where the \textit{min} operator is used to select the minimum bond weight in instances wherein bonds with different weights are matched. We note that as defined by Kretschmer et al \supercite{MCES}, the MCES distance between two molecules does not account for molecular size. While perhaps more qualitatively intuitive, thresholds based on this metric are not practical for assessing the similarity of molecules. This is because the metric does not account for the fact that a bond distance of \textit{d} should be less consequential the greater the number of matched bonds the pair of molecules share. To address this issue, we normalize by the maximum weighted MCES distance possible between two molecules, namely the sum of their respective bond weights. This bounds the MCES distance between 0 and 1. Formally,

\[
d_{\mathrm{MCES^w_{norm}}}(G_1, G_2)
= \frac{d_{\mathrm{MCES^w}(G_1, G_2)}}{\sum_{e \in E_1} b_1(e)
+ \sum_{e \in E_2} b_2(e)
}
\]

\subsubsection*{MCES Dataset}
We used $\mathrm{d_{MCES^w}}$ distances pre-computed by Kretschmer et al to evaluate the ability of molecular representations to capture global topological similarity between molecules. As released, the MCES dataset consists of pairwise distances for 63,437 molecules. Since computing the MCES between two graphs is NP-hard, the fastest known algorithms may take exponential time in the worst case. Thus, as a compromise, Kretschmer et al provide exact distances up to a cut-off of a $\mathrm{d_{MCES^w}}$ of 10. All distances less than 10 are exact values, whereas those greater than or equal to 10 are correct lower bounds on the $\mathrm{d_{MCES^w}}$ distance. Only exact values for pairs with a $\mathrm{d_{MCES^w}} < 10$ were used for the correlation assessment.

Of the 63,437 molecules, 43,443 were sampled from ten commonly-used datasets, namely BACE \supercite{BACE}, BBBP \supercite{BBBP}, ClinTox \supercite{ClinTox}, Delaney \supercite{ESOLDelaney}, Lipo\supercite{Lipo}, SIDER \supercite{SIDER}, Tox21 \supercite{Tox21}, ToxCast \supercite{ToxCast}, SMRT \supercite{SMRT}, and MS/MS \supercite{MCES}. The remaining 19,994 molecules are sampled from the following databases: Kyoto Encyclopedia of Genes and Genomes \supercite{KEGG2025} (KEGG), Chemical Entities of Biological Interest \supercite{ChEBI2026} (ChEBI), Human Metabolome Database \supercite{HMDB2022} (HMDB), Yeast Metabolome Database \supercite{YMDB2017} (YMDB), PlantCyc \supercite{PlantCyc2025}, MetaCyc \supercite{MetaCyc2020}, KNApSAcK \supercite{KNApSAcK2012}, Universal Natural Products Database \supercite{UNPD2013} (UNPD), Mass spectrometry Contaminants Database (MaConDa) \supercite{MaConDa2012}, Hazardous Substances Data Bank (HSDB) \supercite{HSDB2014}, Super Natural II \supercite{SuperNaturalII2015},
COlleCtion of Open Natural prodUcTs (COCONUT) \supercite{COCONUT2025}, and NORMAN \supercite{NORMAN2009}.

Several of the embedding models used in this study impose constraints on the molecules they are able to encode. These constraints include connectivity, atom count, and the type of atoms supported. Moreover, while some models may impose fewer constraints on their inputs, their training datasets often do not include rare elements and disconnected fragments. Thus, to ensure a fair comparison in evaluating graph similarity, SMILES strings were preprocessed to be compliant with the most constrained model. Atom type filtering was performed to remove any molecule featuring any atom not in the set $\{C, O, N, Br, Cl, I, F, S\}$. Since the MCES dataset contains large and inorganic structures, molecules with a molecular weight $\geq$ 900 Da were stripped, as were molecules containing no carbon atoms. Finally, molecules failing to featurize due to token limits were also stripped.

\subsection*{Clustering}

We quantified the frequency of adversarial examples of structurally similar molecules in clustering applications using Butina clustering. Some pairwise calculations in the MCES dataset are censored, as they were not computed. We therefore selected the subset of 19,994 molecules for which all pairwise calculations were computed. This enables the use of the Butina algorithm, which requires a complete pairwise distance matrix. We clustered the MCES dataset using the standard implementation of Butina clustering from the RDKit.ML.Cluster package. As ECFP4 is the field-default for clustering applications, we first featurized all molecules in the MCES dataset using this representation. Butina clustering was conducted with thresholds of 0.4, 0.5, 0.6, 0.7, and 0.8.
\subsection*{Shape similarity}

Using 30,000 molecules randomly sampled from ZINC \supercite{ZINC}, we constructed pairwise similarity matrices for the 12 molecular representations under study. For each representation, we performed equal-width stratified random sampling across the observed similarity range, selecting up to 100 molecular pairs from each of 10 bins, yielding a maximum of 1,000 pairs per representation. For each sampled molecule, we attempted to generate 10 three-dimensional conformers using RDKit. For each molecular pair, we calculated the maximum shape and color similarity scores across all available conformer combinations, optimising the two scores independently using RDKit's AlignMol function. Finally, we calculated Spearman rank correlations between each representation's pairwise similarities and the corresponding shape and color similarity scores, excluding pairs for which valid scores could not be obtained.

\subsection*{ADMET similarity}

To assess whether property landscape roughness correlates with cross-validated model performance on the Biogen dataset, we performed 5-fold cross-validation using generic Bemis Murcko scaffold splits. Property landscape roughness was computed for the full dataset across representations, and this score was correlated with regression performance (mean 5-fold RMSE) across each model type (kNN, random forest, SVR, and a gradient boosted regressor). Note that the 5-fold cross validation was performed without hyperparameter tuning, as our objective was to assess the association between property roughness and out-of-the-box performance of molecular representations. 

The ROGI-XD metric \supercite{ROGIXD} was used to quantify property landscape roughness. ROGI-XD is a scalar metric that quantifies the roughness of a molecular property landscape by measuring how quickly property variation is lost as molecular space is progressively coarse-grained. Given molecular representations $x_i$, property values $y_i$, and a distance metric $d(x_i,x_j)$, molecules are grouped using complete-linkage hierarchical clustering. At each clustering stage, the property value of every molecule in a cluster $C_j$ is replaced by the cluster mean:

$$
\bar y_j=\frac{1}{|C_j|}\sum_{i\in C_j}y_i.
$$

The standard deviation of the resulting coarse-grained property values, denoted by $\sigma_u$, is then calculated. The degree of coarse-graining is defined as

$$
u=1-\frac{\log N_{\mathrm{clusters}}}{\log N},
$$

where $N$ is the total number of molecules. Thus, $u=0$ corresponds to every molecule forming its own cluster, whereas $u=1$ corresponds to all molecules being merged into a single cluster. The ROGI-XD score is defined as the area under the reduction-in-variation curve:

$$
\int_0^1 2\left(\sigma_0-\sigma_u\right) du
$$

where $\sigma_0$ is the standard deviation of the original property values. A high ROGI-XD indicates that substantial property variation is removed when relatively similar molecules are merged, suggesting a rough landscape with frequent activity cliffs, whereas a low value indicates that nearby molecules generally have similar properties and therefore form a smoother landscape. ROGI-XD is particularly suitable for comparing molecular representations because it measures coarse-graining through the fraction of the clustering hierarchy traversed rather than through raw distance thresholds. Raw distances can have very different scales and distributions across fingerprints, descriptor sets, and learned embeddings, especially when their dimensionalities differ. By using the normalized variable $u$, ROGI-XD reduces the effects of distance scaling, distance concentration, and representation-specific distance ranges, allowing a more meaningful comparison of how effectively each representation places molecules with similar property values near one another.

\subsection*{Acknowledgements}

This work was supported by the Biotechnology and Biological Sciences Research Council (BBSRC) and the Reuben Foundation.

\subsection*{Ethics declaration}
The authors declare no competing interests.

\subsection*{Code and data availability }
The code and data for this work are available at the following repository: \url{https://github.com/oxpig/On-the-limitations-of-molecular-representations}.

\subsection*{Author contributions} 

A.B, F.I, and C.M.D designed the study. A.B conducted the experiments and analysed the results. F.I and C.M.D supervised the project. All authors discussed the results and contributed to the interpretation. A.B wrote the manuscript with editorial input from all authors.

\printbibliography

\begin{extendedfigure*}[p]
        \rule{\linewidth}{1.2pt}
        \centering
        
        \vspace{-0.5cm}

        \makebox[\textwidth][c]{%
            \hspace*{-0.8cm}%
            \includegraphics[scale=1.4]{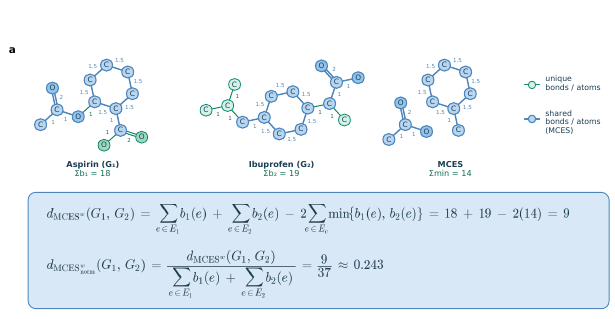}%
        }
        
        \vspace{0.5cm}
        \includegraphics[
            width=\textwidth,
            height=0.85\textheight,
            keepaspectratio
        ]{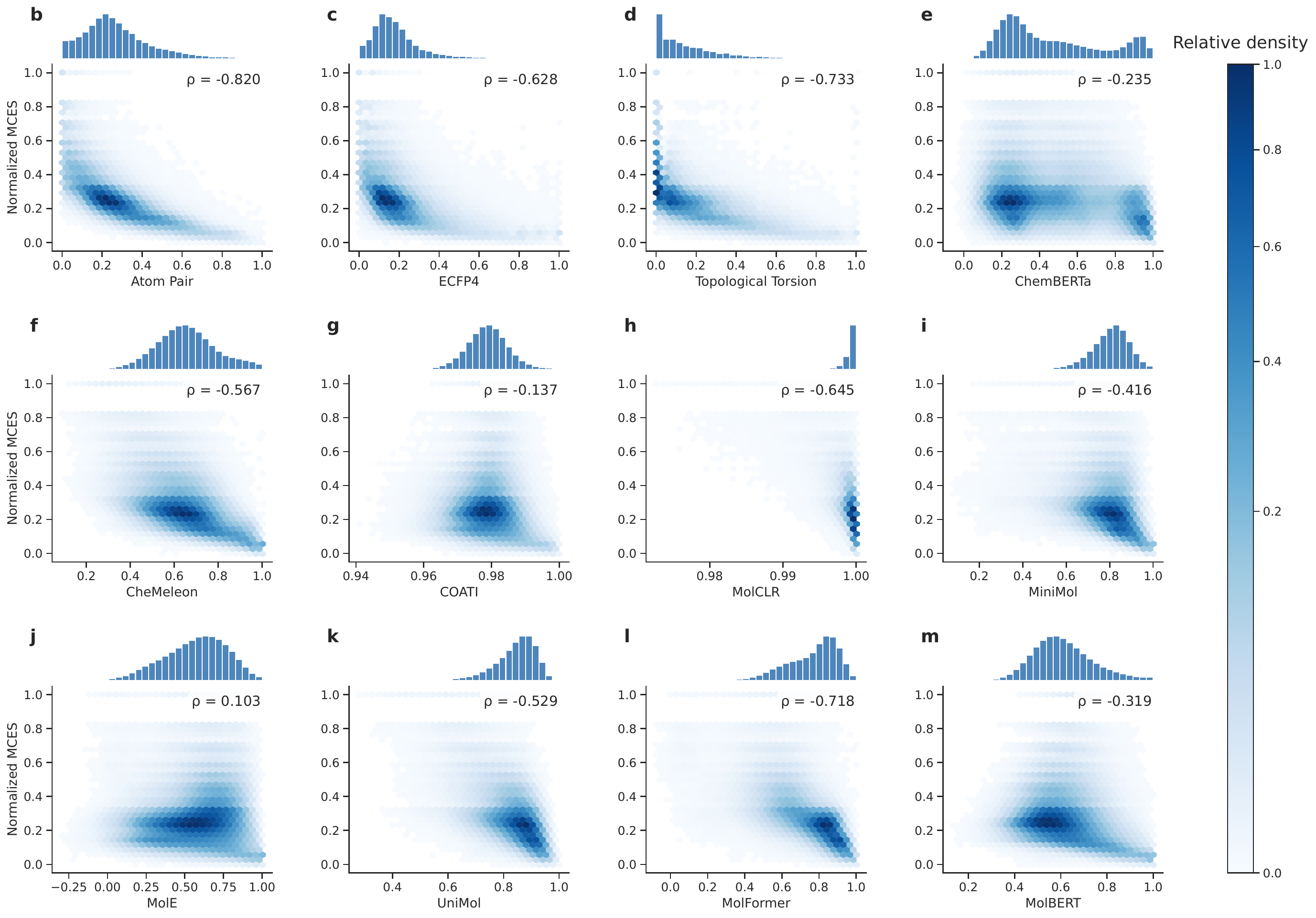}

    \caption{\textbf{Graph similarity evaluations}. \textbf{(a)} An example of normalized weighted MCES distance between two molecules. \textbf{(b-m)} Hexbin relative density plots with marginal distributions showing the Spearman correlation between Tanimoto similarity \textbf{(b-d)} and cosine similarity \textbf{(e-m)} against normalized weighted MCES distance across 12 representations. After preprocessing, a total of 4,348,627 pairwise comparisons are plotted for each hexbin plot. All $d_{MCES^w}$ values are $< 10$.}
    \rule{\linewidth}{1.2pt}
    \label{fig:mces-hexbin}
    \phantomsubcaption\label{fig:mces-hexbin-a}
    \phantomsubcaption\label{fig:mces-hexbin-b}
    \phantomsubcaption\label{fig:mces-hexbin-c}
    \phantomsubcaption\label{fig:mces-hexbin-d}
    \phantomsubcaption\label{fig:mces-hexbin-e}
    \phantomsubcaption\label{fig:mces-hexbin-f}
    \phantomsubcaption\label{fig:mces-hexbin-g}
    \phantomsubcaption\label{fig:mces-hexbin-h}
    \phantomsubcaption\label{fig:mces-hexbin-i}
    \phantomsubcaption\label{fig:mces-hexbin-j}
    \phantomsubcaption\label{fig:mces-hexbin-k}
    \phantomsubcaption\label{fig:mces-hexbin-l}
    \phantomsubcaption\label{fig:mces-hexbin-m}
\end{extendedfigure*}

\begin{extendedfigure*}[p]
  \rule{\linewidth}{1.2pt}
    \centering
    \hspace*{-0.5cm}
    \includegraphics[scale=0.475]{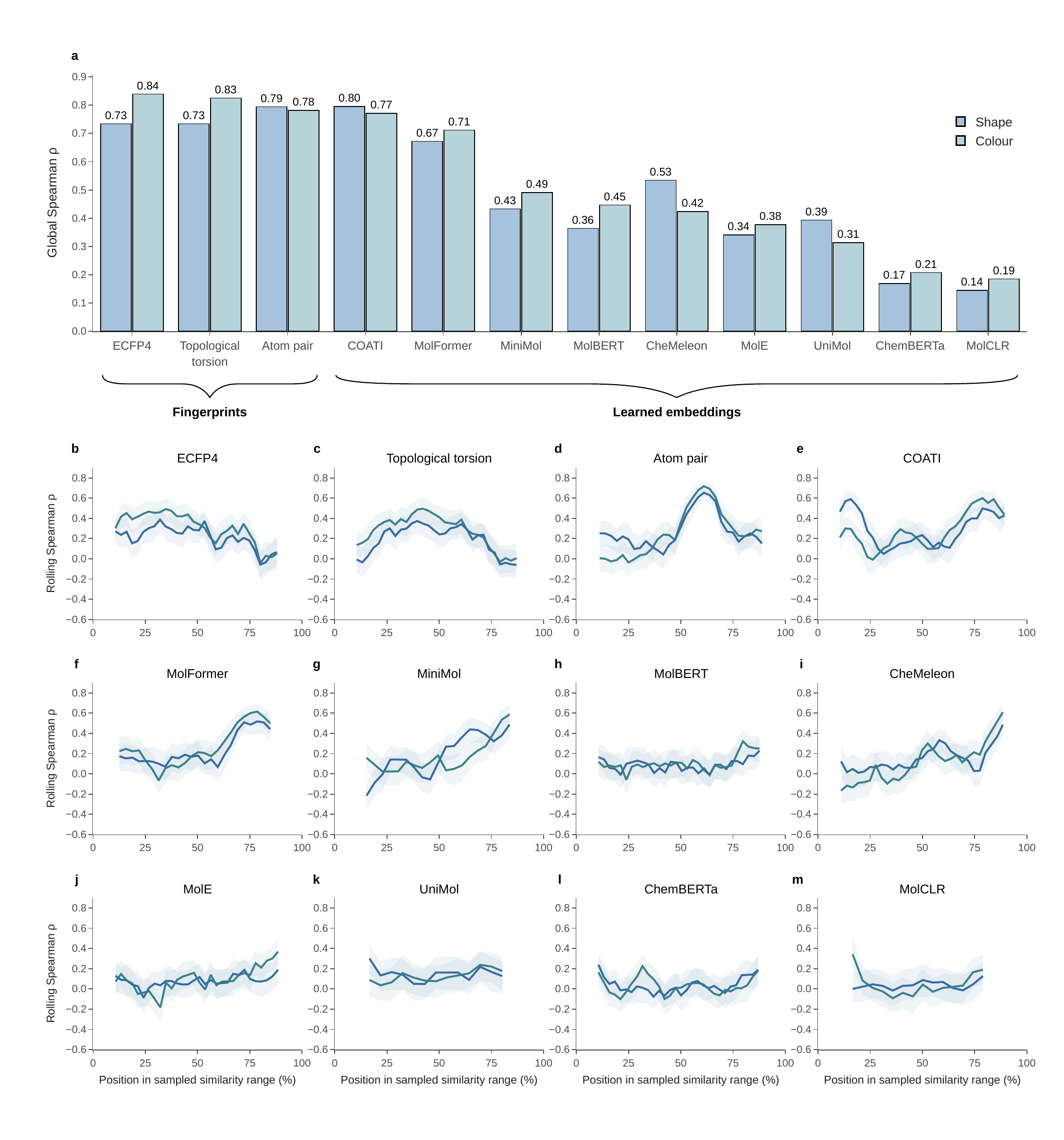}
    
    \vspace{-4mm}
    \caption{\textbf{Shape evaluations for molecular representations}. \textbf{(a)} Bar chart showing the Spearman correlation between Tanimoto and cosine similarity against maximum pairwise conformational overlap score for randomly sampled molecules from ZINC. Overlap scores are given by RDKit shape and color score returned by the RDKit AlignMol API. For each representation, 1,000 total molecular pairs were uniformly sampled according to the given model's similarity distribution. Each model's distribution was divided into ten equal-width intervals and up to 100 pairs were randomly sampled per similarity bin. 10 conformers were generated per molecule and the maximum pairwise score was taken for each pair. \textbf{(b-m)} Rolling shape and color Spearman correlations for each of the 12 representations across binned (200-molecular pairs) intervals with a step size of 25 and shaded regions representing 95\% confidence intervals estimated from 300 bootstrap resamples with replacement.}
    \vspace*{-0.15cm}
    \rule{\linewidth}{1.2pt}
    \label{fig:shape-evaluations}
    \phantomsubcaption\label{fig:shape-evaluations-a}
    \phantomsubcaption\label{fig:shape-evaluations-b}
    \phantomsubcaption\label{fig:shape-evaluations-c}
    \phantomsubcaption\label{fig:shape-evaluations-d}
    \phantomsubcaption\label{fig:shape-evaluations-e}
    \phantomsubcaption\label{fig:shape-evaluations-f}
    \phantomsubcaption\label{fig:shape-evaluations-g}
    \phantomsubcaption\label{fig:shape-evaluations-h}
    \phantomsubcaption\label{fig:shape-evaluations-i}
    \phantomsubcaption\label{fig:shape-evaluations-j}
    \phantomsubcaption\label{fig:shape-evaluations-k}
    \phantomsubcaption\label{fig:shape-evaluations-l}
    \phantomsubcaption\label{fig:shape-evaluations-m}
\end{extendedfigure*}

\begin{extendedfigure*}[p]
    \rule{\linewidth}{1.2pt}
    \centering

    \makebox[\linewidth][c]{%
        \includegraphics[
            width=1.175\linewidth,
            trim=0 110 0 0,
            clip
        ]{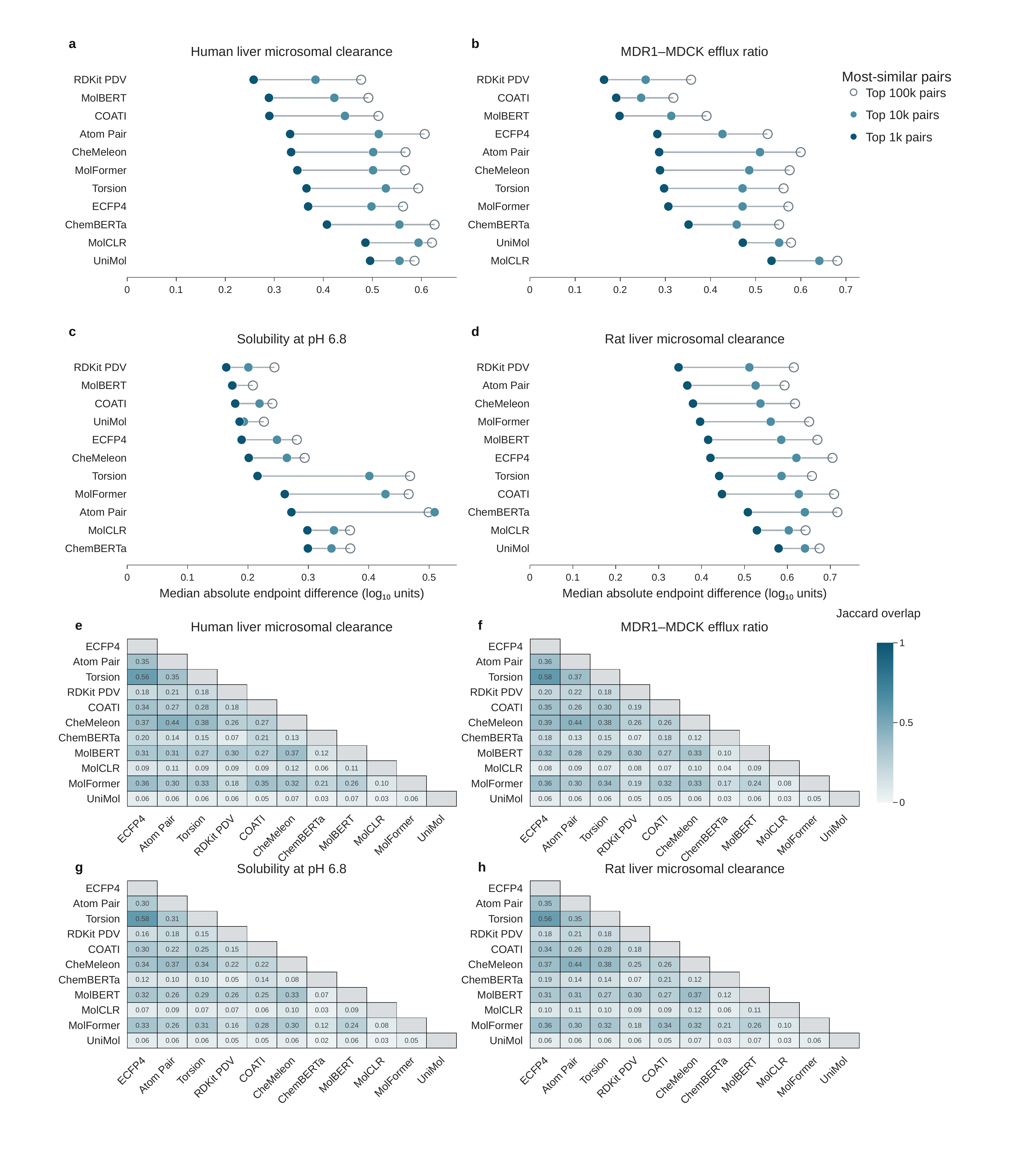}%
    }

    \vspace{-2mm}

    \caption{\textbf{Chemical similarity evaluations for molecular representations}. \textbf{(a-d)} ADMET endpoint similarity for the Biogen dataset across similarity thresholds. Each plot shows the median absolute endpoint difference for the top closest 100,000, 10,000 and 1,000 molecular pairs by representation similarity. \textbf{(e-h)} Jaccard similarity triangles indicating the fraction of molecules that overlap in the top 1,000 of molecular pairs by representation similarity. }
    \label{fig:chem-evals}
    \phantomsubcaption\label{fig:chem-evals-a}
    \phantomsubcaption\label{fig:chem-evals-b}
    \phantomsubcaption\label{fig:chem-evals-c}
    \phantomsubcaption\label{fig:chem-evals-d}
    \phantomsubcaption\label{fig:chem-evals-e}
    \phantomsubcaption\label{fig:chem-evals-f}
    \phantomsubcaption\label{fig:chem-evals-g}
    \phantomsubcaption\label{fig:chem-evals-h}

    \rule{\linewidth}{1.2pt}
\end{extendedfigure*}

\begin{extendedfigure*}[!t]

    \rule{\linewidth}{1.2pt}

    \hspace*{-0.05\linewidth}%
    \begin{subfigure}[b]{1.075\linewidth}
        \centering
        \includegraphics[
            width=\linewidth,
            trim={0 24mm 0 15mm},
            clip
        ]{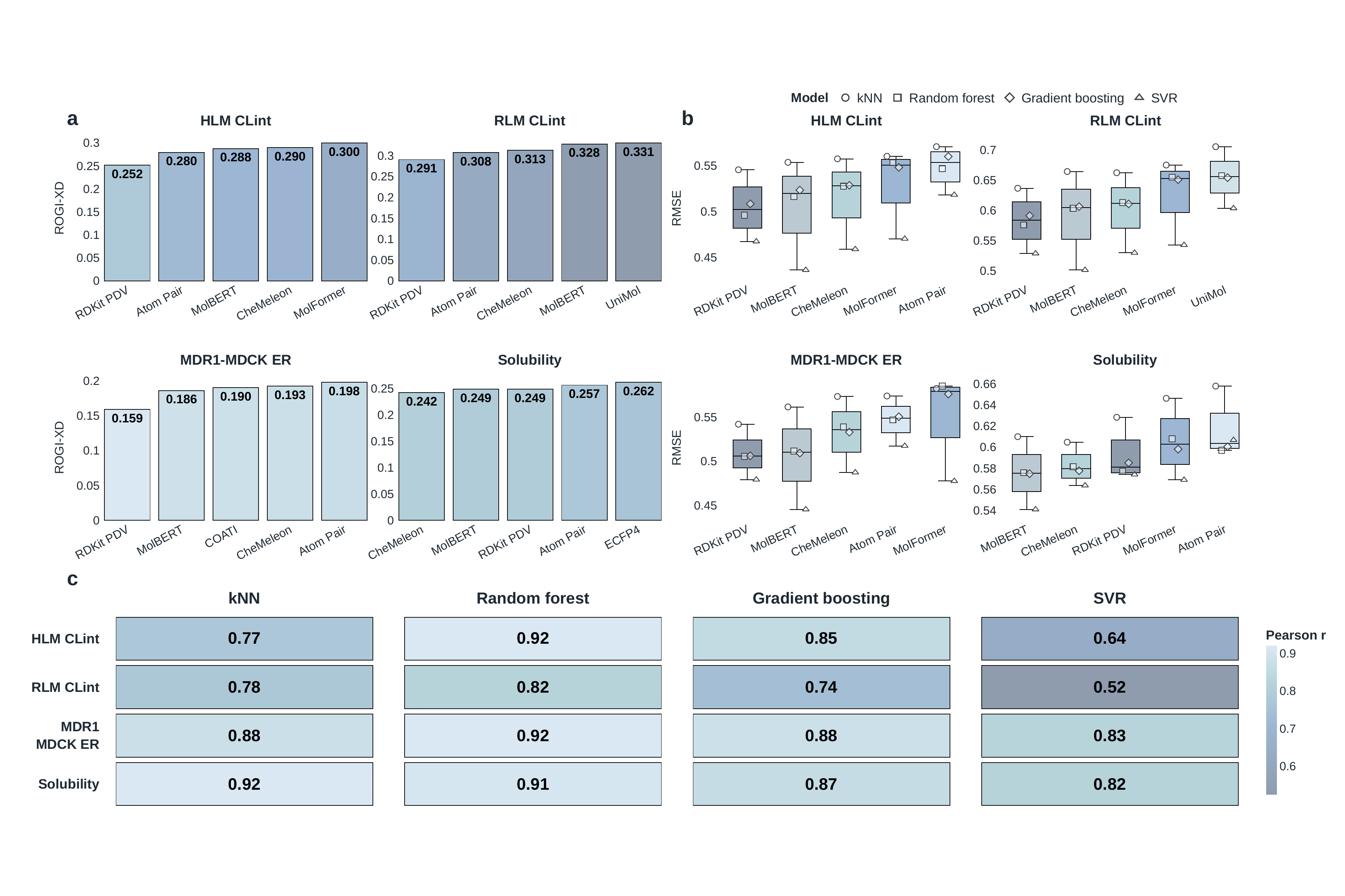}
    \phantomsubcaption\label{fig:rogi-evals-a}
    \phantomsubcaption\label{fig:rogi-evals-b}
    \phantomsubcaption\label{fig:rogi-evals-c}
    \label{fig:rogi-evals}
    \end{subfigure}

    \caption{\textbf{QSAR roughness and performance}. \textbf{(a)}  ROGI-XD roughness for the representations with the lowest ROGI-XD values. ROGI-XD is computed across all molecules for which endpoint measurements are available. Lower values indicate representations that induce a smoother landscape. \textbf{(b)} Box plots for mean 5-fold RMSE for the top 5 representations across each endpoint and model. \textbf{(c)} Pearson correlation between ROGI-XD roughness and cross-validated RMSE. }

    \vspace{-0.15cm}
    \rule{\linewidth}{1.2pt}
\end{extendedfigure*}


\end{document}


\maketitle

\section{Supplementary Figures}

\begin{figure}[H]
  \rule{\linewidth}{1.2pt}\par
  \vspace{0.5cm}
    \centering
    \includegraphics[scale=0.4]{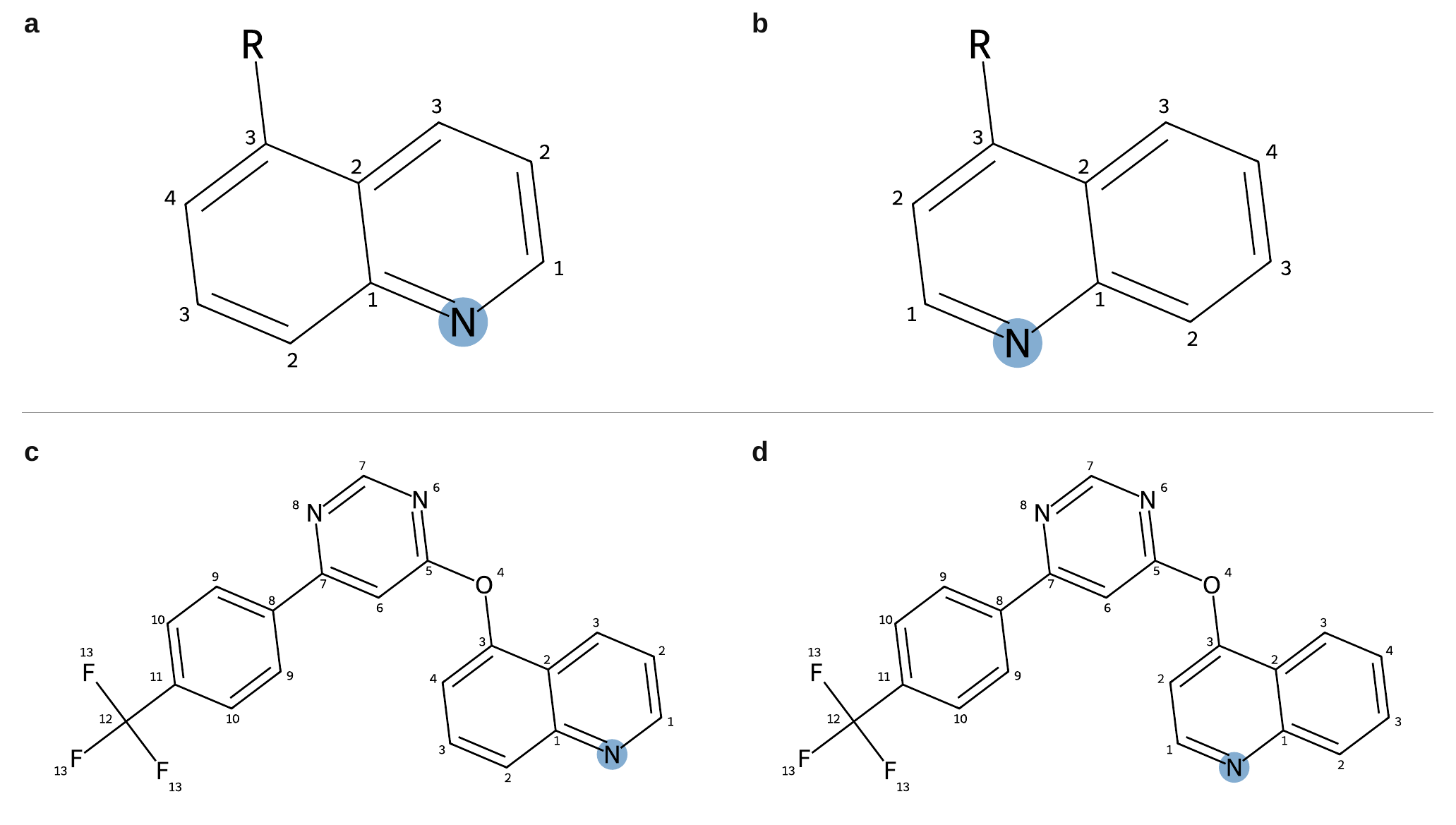}
    \caption{\textbf{Quinoline inversion construction for atom pair fingerprints.} The figure shows how the nitrogen atom in a quinoline functional group \textbf{(a)} can be relocated to the same position on the opposite fused ring \textbf{(b)} while maintaining fingerprint identity. R groups here signify that provided the same structure is substituted for both placeholders in subfigures \textbf{(a)} and \textbf{(b)}, the resulting molecule will have the same fingerprint. Numerical annotations represent the bond distance from the highlighted nitrogen to each atom. Atom pair fingerprints encode features using a three-tuple of the form $(a_1, d, a_2)$ where $a_1$ and $a_2$ are the atom environments of two distinct atoms and $d$ is the bond distance between them.  Atom environments are encoded using atomic number, number of $\pi$ electrons, and degree. None of these values are changed by the relocation of the nitrogen. Further, because of the symmetry of the quinoline functional group, bond distances are matched by the relocation of the nitrogen. Subfigures \textbf{(c)} and \textbf{(d)} show an example pair of degenerate molecules satisfying the rule depicted.}
    \label{fig:quinoline inversion}
    \rule{\linewidth}{1.2pt}
\end{figure}

\newpage
\begin{figure}[H]
\vspace{5cm}
    \centering
    \rule{\linewidth}{1.2pt}\par
    \vspace{0.5cm}
    \includegraphics[scale=0.3]{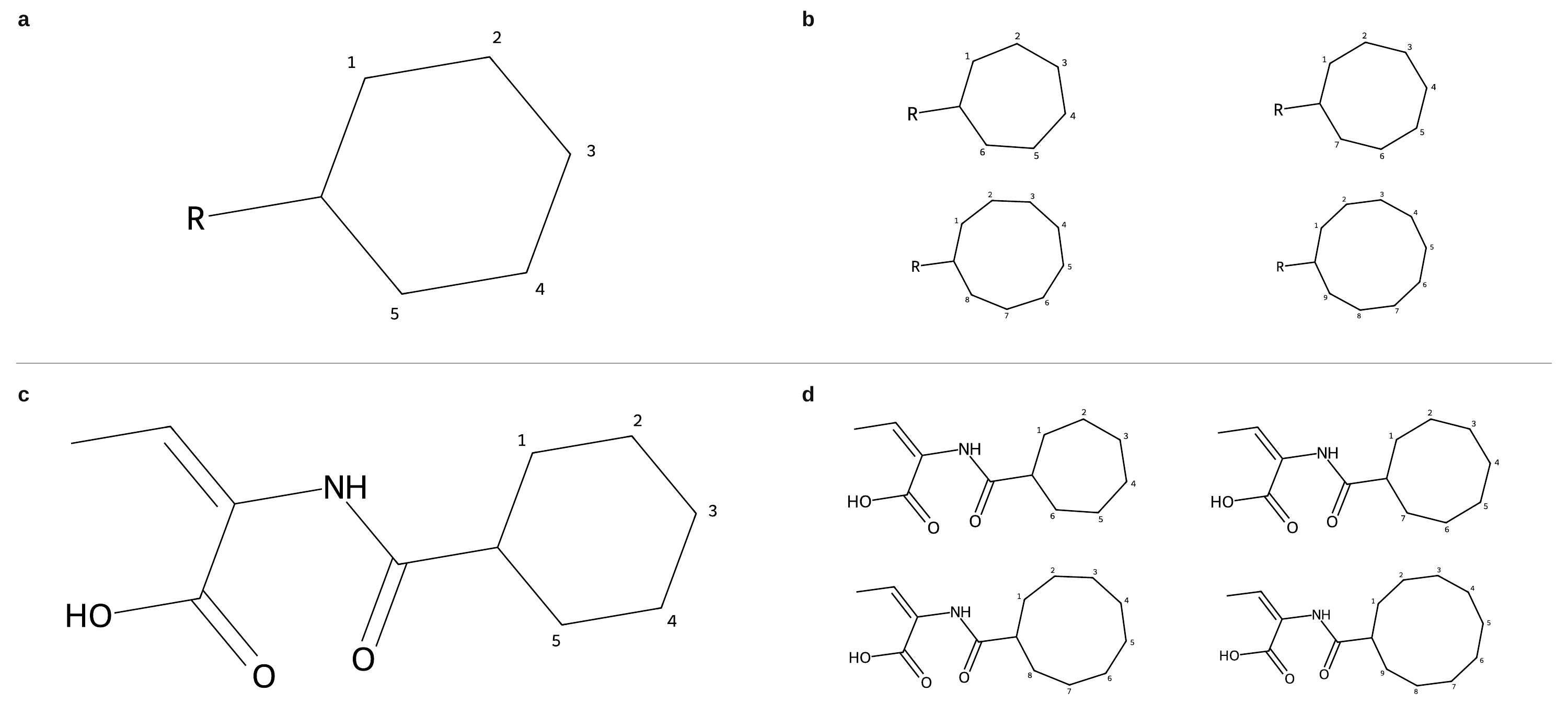}
    \caption{\textbf{Ring expansion for ECFP4 fingerprints.} The figure shows how non-aromatic rings can be expanded while maintaining fingerprint identity for ECFP4 fingerprints. R groups here signify that provided the same structure is substituted for both placeholders in subfigures \textbf{(a)} and \textbf{(b)}, the resulting molecule will have the same fingerprint. Note that there is no limit as to how large the rings can be made in the cases of subfigures \textbf{(b)} and \textbf{(d)}. The maintenance of fingerprint identity can be attributed to the fact that no new radial substructures up to a bond distance of 2 are generated. Note that it is not necessary for rings to be cycloalkanes for ring expansion to be possible. It is possible to replace these with heterocycles in some instances.}
    \label{fig:ring expansion}
    \rule{\linewidth}{1.2pt}
\end{figure}

\newpage
\begin{figure}[H]
    \centering
    \vspace{5cm}
    \rule{\linewidth}{1.2pt}\par
    \vspace{0.5cm}
    \includegraphics[scale=0.4]{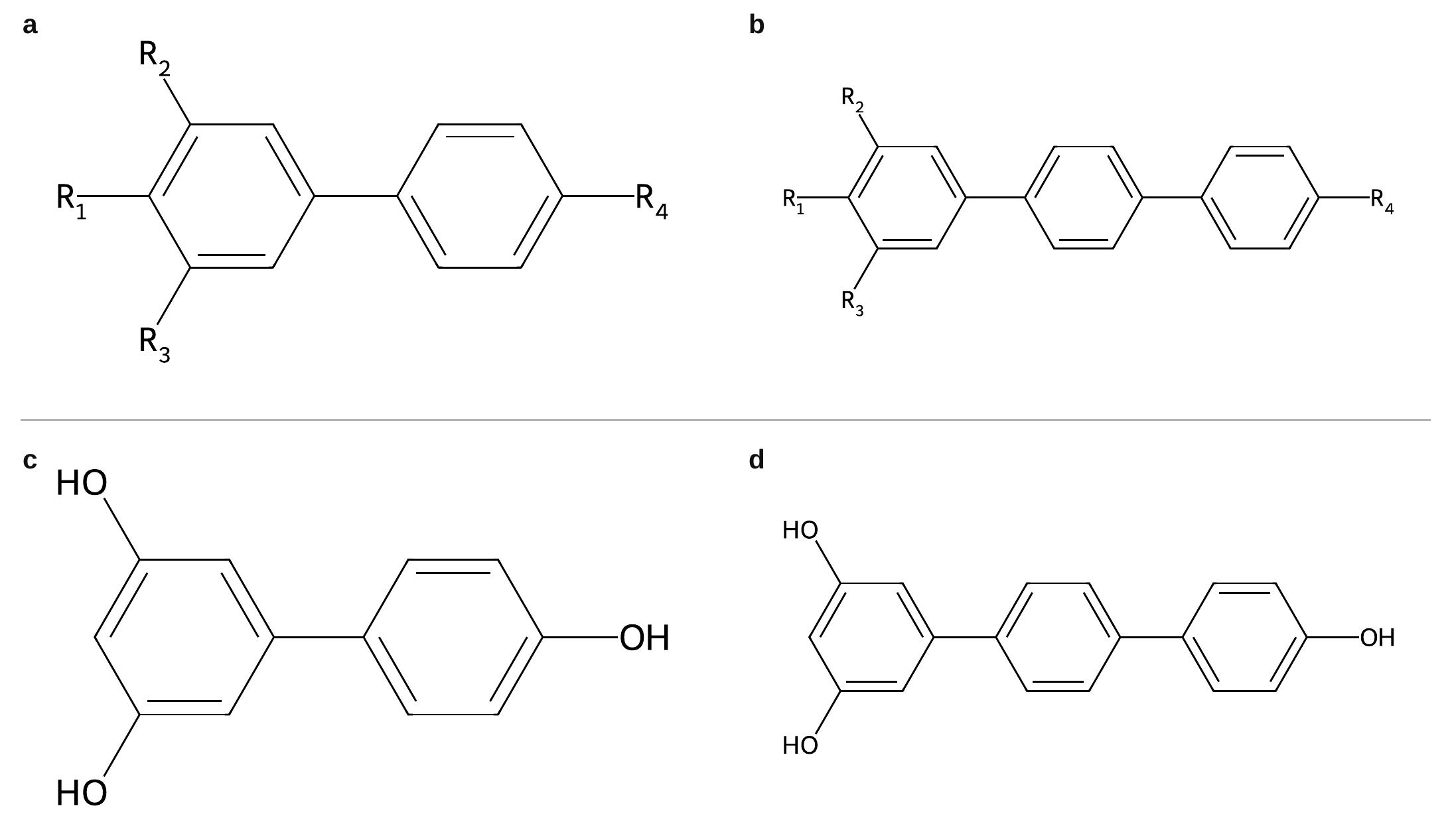}
    \caption{\textbf{Ring insertion for ECFP4 fingerprints.} The figure shows how additional benzene groups can be inserted into a molecular structure while maintaining fingerprint identity. R groups here signify that provided the same structure is substituted for both structures in subfigures \textbf{(a)} and \textbf{(b)}, the resulting molecule will have the same fingerprint. Note that there is no limit as to how many rings may be inserted into the structures of \textbf{(b)} and \textbf{(d)}. The maintenance of fingerprint identity can be attributed to the fact that no new radial substructures up to a bond distance of 2 are generated. Ring insertion is a specific case of substructure insertion, wherein molecules degenerate under the ECFP4 fingerprint can be constructed by repeatedly appending repeated substructures to a reference molecule. Provided the attachment points for the repeated substructures do not introduce a new substructure, these molecules will maintain an identical fingerprint.}
    \label{fig:ring insertion}
    \rule{\linewidth}{1.2pt}

\end{figure}

\newpage
\begin{figure}[H]
    \centering
    \vspace{5cm}
    \rule{\linewidth}{1.2pt}\par
    \vspace{0.5cm}
    \includegraphics[scale=0.4]{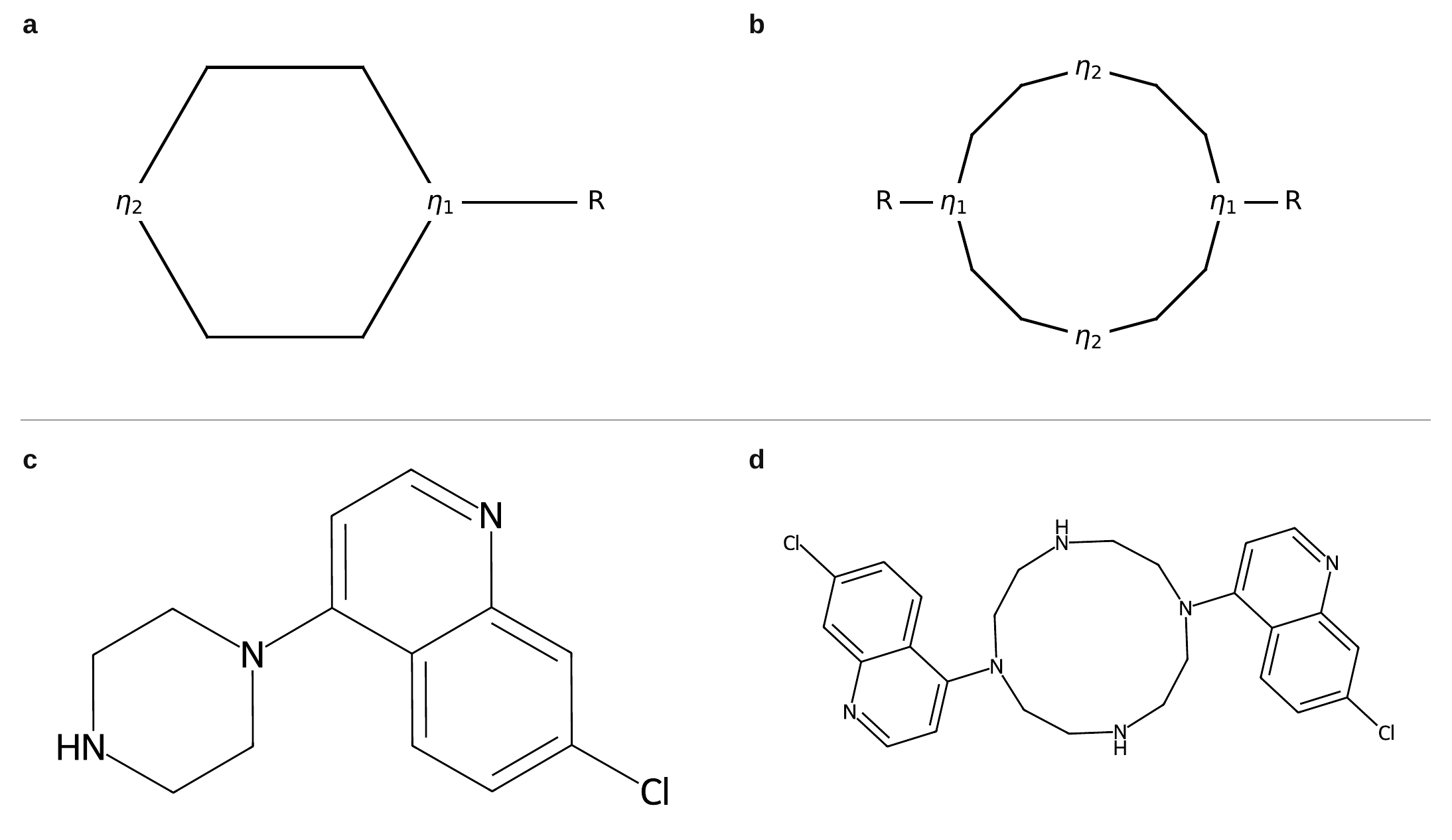}
    \caption{\textbf{Macrocyclization for ECFP4 fingerprints.} The figure shows how a 6-membered heterocycle with heteroatoms at the 1 and 4 positions and an attached side-chain can be resized into a 12-membered macrocycle while maintaining fingerprint identity. R groups here signify that provided the same structure is substituted for the side chains in subfigures \textbf{(a)} and \textbf{(b)}, the resulting molecules (subfigures \textbf{(c)} and \textbf{(d)}) will have the same fingerprint. The atoms denoted $\eta _1$, $\eta _2$ may be any heteroatoms satisfying valence rules. The maintenance of fingerprint identity can be attributed to the fact that the constructed 12-membered macrocycle maintains the same radius 2 substructures as the 6-membered heterocycle throughout. Indeed, provided the period of repeated heteroatoms and side-chains repeats with the same frequency, the heterocycle can be expanded infinitely.}
    \label{fig:macrocyclization}
    \rule{\linewidth}{1.2pt}
\end{figure}

\newpage
\begin{figure}[H]
    \vspace{5cm}
    \centering
    \rule{\linewidth}{1.2pt}\par
    \vspace{0.5cm}
    \includegraphics[scale=0.4]{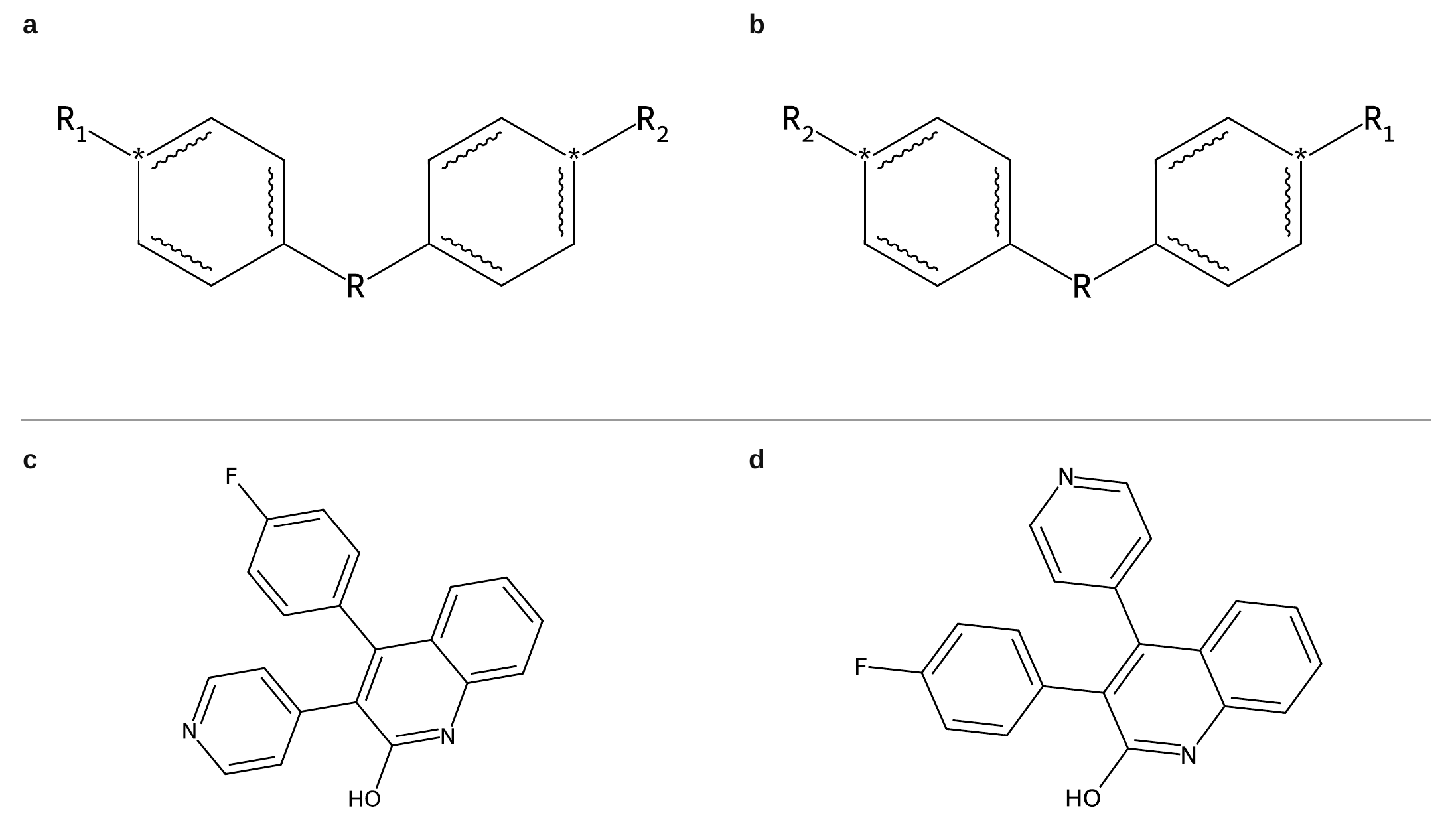}
    \caption{\textbf{Ring swap for ECFP4 and topological torsion fingerprints.} The figure shows how two substituted rings can be exchanged with respect to a shared skeleton while maintaining fingerprint identity. R groups here signify that provided the same structures are substituted for both $R_1$ and $R_2$ in figures \textbf{(a)} and \textbf{(b)}, the resulting molecule will have the same fingerprint. The wavy double bonds indicate that the rings may be either aromatic or saturated. Further, the asterisks indicate that the given atoms may be heteroatoms. The maintenance of identical ECFP4 and topological torsion fingerprints can be attributed to the fact that no new radial substructures are introduced (accounting for ECFP4 identity) nor are any new length 4 sequences of atoms (accounting for topological torsion identity). Note that the construction is not limited to two substituted rings being exchanged. In fact, provided the rings follow the same structure as in \textbf{(a)} and \textbf{(b)} any number of rings can be swapped. Figures \textbf{(c)} and \textbf{(d)} provide a concrete example of ring swaps that maintain fingerprint identity.}
    \label{fig:ring swap}
    \rule{\linewidth}{1.2pt}
\end{figure}

\newpage
\begin{figure}[H]
    \vspace{5cm}
    \centering
    \rule{\linewidth}{1.2pt}\par
    \vspace{0.5cm}
    \includegraphics[scale=0.4]{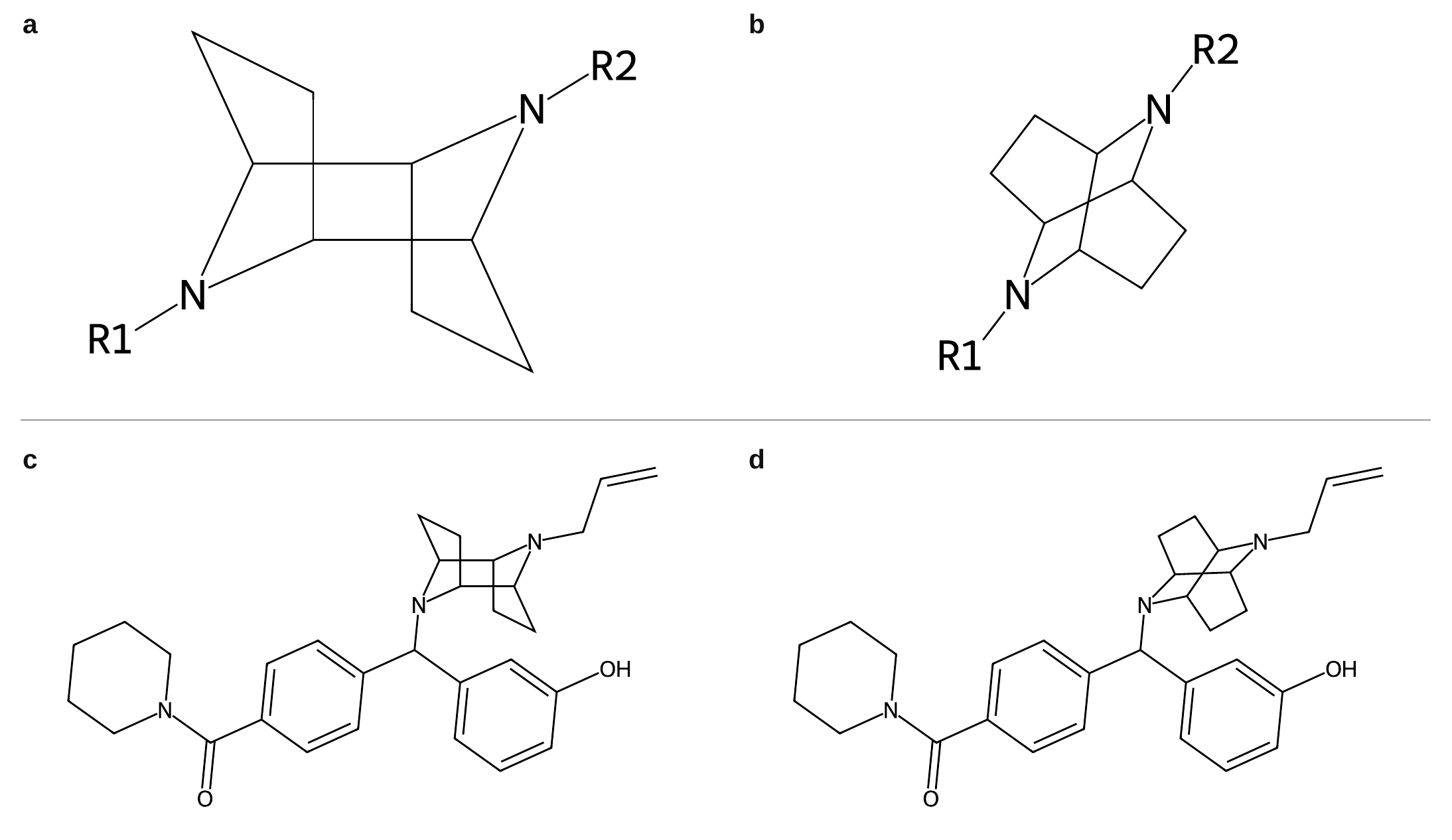}
    \caption{\textbf{Cage substitution for ECFP4, topological torsion and atom pair fingerprints.} The figure depicts how diazatricyclodecane functional groups can be substituted for each other while maintaining fingerprint identity. Subfigure \textbf{(a)} depicts a 9,10-diazatricyclo[4.2.1.1$^2$,$^5$]decane ring, while \textbf{(b)} represents a 2,7-diazatricyclo[4.4.0.0$^3$,$^8$]decane ring. The R groups here signify that provided the same structures are substituted for both $R_1$ and $R_2$ in figures a and b, the resulting molecule will have the same fingerprint. Figures (\textbf{c}) and (\textbf{d}) provide a concrete example of two molecules sharing the same fingerprint with substituted diazatricyclodecane groups.}
    \label{fig:cage substitution}
    \rule{\linewidth}{1.2pt}
\end{figure}

\begin{figure}[H]
    \vspace{5cm}
    \centering
    \rule{\linewidth}{1.2pt}\par
    \vspace{0.5cm}
    \includegraphics[scale=0.4]{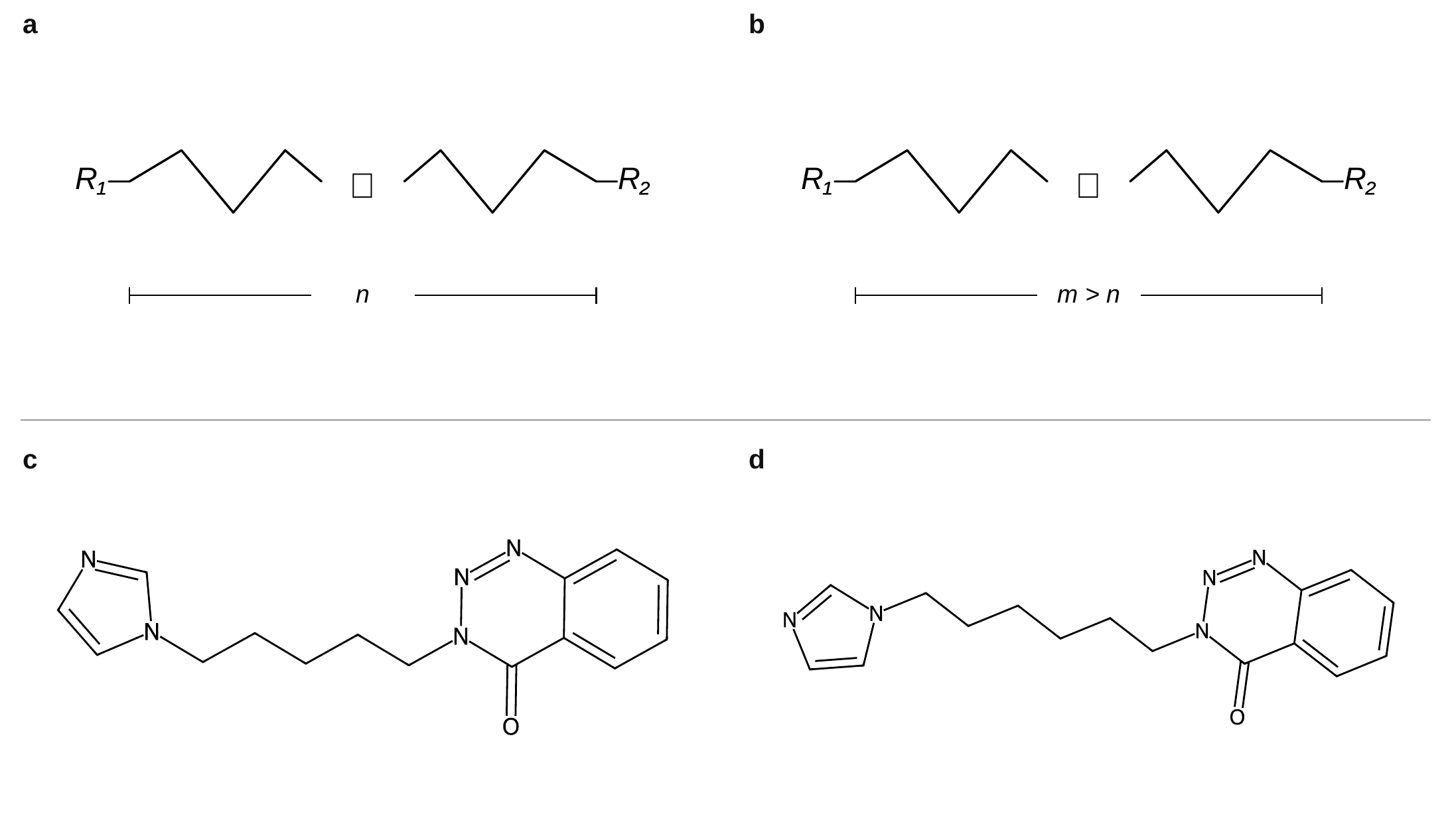}
    \caption{\textbf{Linker extension for ECFP4 and topological torsion fingerprints.} The figure shows how a carbon chain can be extended in length while maintaining fingerprint identity. R groups here signify that provided the same structure is substituted for both structures in figures \textbf{(a)} and \textbf{(b)}, the resulting molecule will have the same fingerprint. Note that the parity of the length of the original linker may be required to be equivalent for degeneracy to hold. Figures (\textbf{c}) and (\textbf{d}) provide a concrete example of two molecules sharing the same fingerprint with carbon-chain linkers of varying lengths.}
    \label{fig:linker extension}
    \rule{\linewidth}{1.2pt}
\end{figure}

\begin{figure}[H]
    \vspace{5cm}
    \centering
    \rule{\linewidth}{1.2pt}\par
    \vspace{0.5cm}
    \includegraphics[scale=0.4]{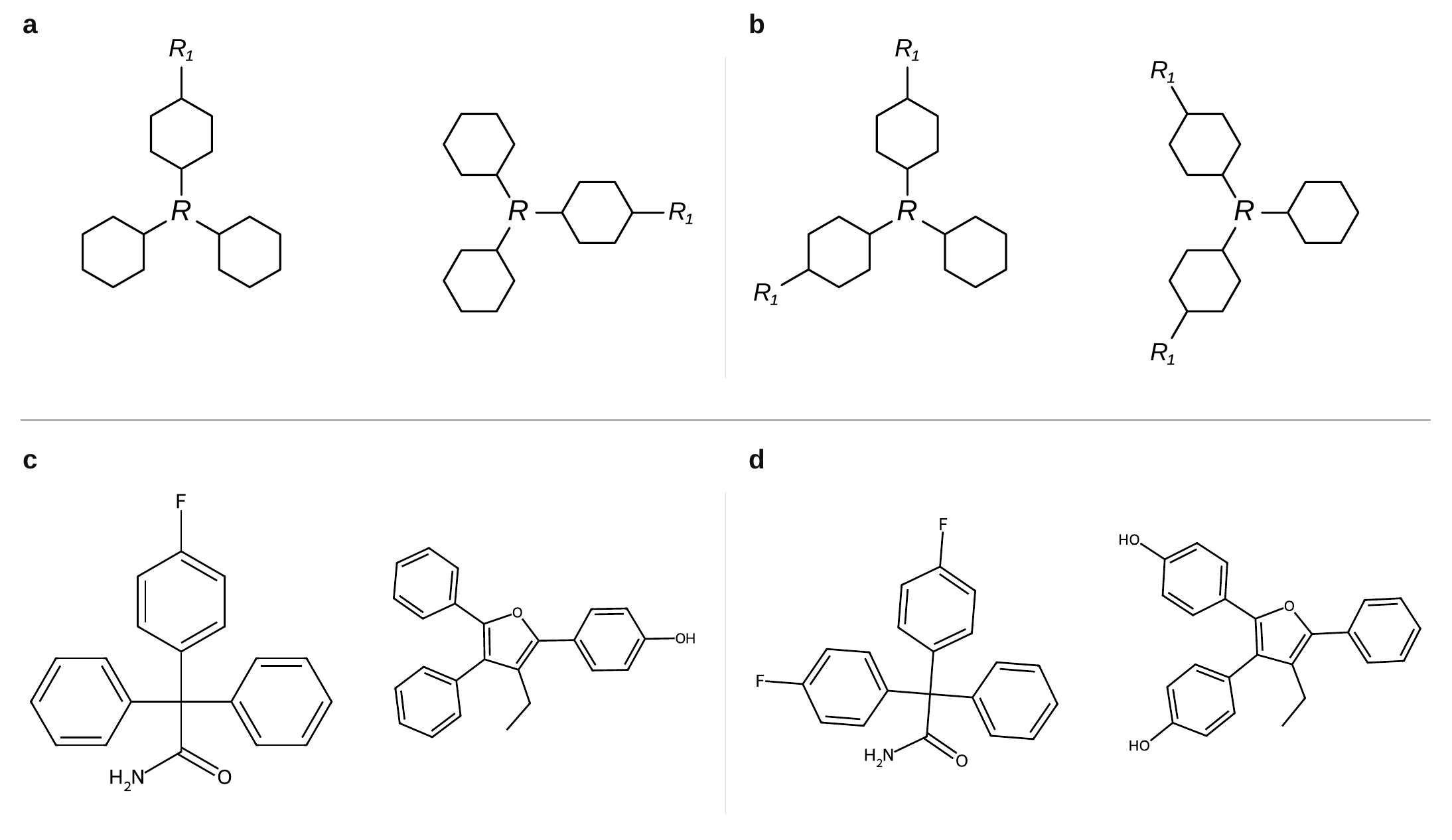}
    \caption{\textbf{Substituent swap and duplication for ECFP4 fingerprints.} The figure shows how ring substituents can be relocated and duplicated on unsubstituted rings within the same molecule while maintaining fingerprint identity. R groups here signify that provided the same structure is substituted for the corresponding structures in subfigures \textbf{(a)} and \textbf{(b)}, the resulting molecule will have the same fingerprint. Figures (\textbf{c}) and (\textbf{d}) provide two examples of substituent swaps and duplications for molecules that maintain fingerprint identity.}
    \label{fig:quinoline inversion}
    \rule{\linewidth}{1.2pt}
\end{figure}